\documentclass{article}

\usepackage{arxiv1}

\usepackage{url,hyperref,lineno,microtype,subcaption}

\usepackage{graphicx,soul}
\usepackage{amsmath}
\usepackage{bm}
\usepackage{epstopdf}
\usepackage{epsfig}
\usepackage{subcaption}
\usepackage{caption}
\usepackage{hyperref}
\usepackage{xcolor}
\usepackage{soul}
\usepackage{bbm}
\usepackage{wrapfig}
\usepackage{floatrow}
\usepackage{siunitx}

\usepackage{xr}

\usepackage[utf8]{inputenc} % allow utf-8 input
\usepackage[T1]{fontenc}    % use 8-bit T1 fonts
\usepackage{hyperref}       % hyperlinks
\usepackage{url}            % simple URL typesetting
\usepackage{booktabs}       % professional-quality tables
\usepackage{amsfonts}       % blackboard math symbols
\usepackage{nicefrac}       % compact symbols for 1/2, etc.
\usepackage{microtype}      % microtypography
\usepackage{lipsum}
\usepackage{graphicx}
\graphicspath{ {./images/} }

\usepackage{ulem}
\usepackage{color}
\newcommand{\new} [1]{\textcolor{black}{{#1}}}

\title{Exploiting multiple timescales in hierarchical echo state networks}

\author{
 Luca Manneschi \\
  Department of Computer Science\\
  University of Sheffield\\
  Sheffield, UK, \\
  \small{\texttt{lmanneschi1@sheffield.ac.uk}} \\
   \And
 Matthew O. A. Ellis \\
  Department of Computer Science\\
  University of Sheffield\\
  Sheffield, UK, \\
  \small{\texttt{m.o.ellis@sheffield.ac.uk }} \\
  \And
 Guido Gicante \\
  National Center for Radiation Protection \\ and Computational Physics,\\
  Italian Institute of Health\\
  Rome, IT \\
  \small{\texttt{guido.gigante@gmail.com}} \\
  \And
 Andrew C. Lin \\
  Department of Biomedical \\Science, \\ 
  The University of Sheffield, \\ 
  Sheffield, UK,\\
  \small{\texttt{andrew.lin@sheffield.ac.uk}} \\
  \And
Paolo Del Giudice$^\dagger$ \\
  National Center for Radiation Protection \\ and Computational Physics,\\
  Italian Institute of Health\\
  Rome, IT, \\
  \small{\texttt{paolo.delgiudice@iss.it}} \\
  \And
Eleni Vasilaki$^\dagger$ \\
Department of Computer Science\\
  University of Sheffield\\
  Sheffield, UK, \\
  \small{\texttt{e.vasilaki@sheffield.ac.uk}} \\ \\
  $^{\dagger}$Joint senior authorship
}

\begin{document}
\maketitle

\begin{abstract}
Echo state networks (ESNs) are a powerful form of reservoir computing that only require training of linear output weights whilst the internal reservoir is formed of fixed randomly connected neurons. With a correctly scaled connectivity matrix, the neurons' activity exhibits the echo-state property and responds to the input dynamics with certain timescales. Tuning the timescales of the network can be necessary for treating certain tasks, and some environments require multiple timescales for an efficient representation. Here we explore the timescales in hierarchical ESNs, where the reservoir is partitioned into two smaller linked reservoirs with distinct properties. Over three different tasks (NARMA10, a reconstruction task in a volatile environment, and psMNIST), we show that by selecting the hyper-parameters of each partition such that they focus on different timescales, we achieve a significant performance improvement over a single ESN. Through a linear analysis, and under the assumption that the timescales of the first partition are much shorter than the second's (typically corresponding to optimal operating conditions), we interpret the feedforward coupling of the partitions in terms of an effective representation of the input signal, provided by the first partition to the second, whereby the instantaneous input signal is expanded into a weighted combination of its time derivatives. 
Furthermore, we propose a data-driven approach to optimise the hyper-parameters through a gradient descent optimisation method that is an online approximation of backpropagation through time. We demonstrate the application of the online learning rule across all the tasks considered.  
\end{abstract}

% keywords can be removed

\section{Introduction}
The high inter-connectivity and asynchronous loop structure of Recurrent Neural Networks (RNNs) make them powerful techniques for processing temporal signals\cite{Ludik1997}. 
%Due to the complex inter-connectivity of RNNs, they cannot be trained using the conventional back-propagation algorithm used in feed forward networks. This arise due to dependence of each neurons state on other neurons at previous times. 
However, the complex inter-connectivity of RNNs means that they cannot be trained using the conventional back-propagation (BP) algorithm\cite{Rumelhart1985learning} used in feed-forward networks, since each neuron's state depends on other neuronal activities at previous times.
A method known as Back-Propagation-Through-Time (BPTT) \cite{Werbos1990}, which relies on an unrolling of neurons' connectivity through time to propagate the error signal to earlier time states, can be prohibitively complex for large networks or time series. Moreover, BPTT is not considered biologically
plausible as neurons must retain memory of their activation over the length of the input and the error signal must be propagated backwards with symmetric synaptic weights \cite{bellec2020solution}.

Many of these problems can be avoided using an alternative approach: reservoir computing (RC). In the subset of RC networks known as Echo State networks, a fixed `reservoir' transforms a temporal input signal in such a way that only a single layer output perceptron needs to be trained to solve a learning task. %\pdg{It seems to me that this definition, as you write later, fits the Echo State networks, RC is more general, and can include plasticity in the reservoir; I would just say something like `in the subset of RC networks known as Echo State networks, ...', or stay general here and later restrict to ESN, as in lines 68-69}. 
The advantage of RC is that the reservoir is a fixed system that can be either computationally or physically defined. Since it is fixed it is not necessary to train the reservoir parameters through BPTT, making RC networks much simpler to train than RNNs. Furthermore, the random structure of a RC network renders the input history over widely different time-scales, offering a representation that can be used for a wide variety of tasks without optimising the recurrent connectivity between nodes. 

Reservoirs have biological analogues in cerebellum-like networks (such as the cerebellum, the insect mushroom body and the electrosensory lobe of electric fish), in which input signals encoded by relatively few neurons are transformed via `expansion re-coding' into a higher-dimensional space in the next layer of the network, which has many more neurons than the input layer \cite{Marr1969theory, Farris2011mushroom, Laurent2002olfactory, Warren2016comparative}. This large population of neurons (granule cells in the cerebellum; Kenyon cells in the mushroom body) acts as a reservoir because their input connectivity is fixed and learning occurs only at their output synapses. The principal neurons of the `reservoir' can form chemical and electrical synapses on each other (e.g. Kenyon cells: \cite{takemura2017connectome, zheng2018complete, liu2016gap}), analogous to the recurrent connectivity in reservoir computing that allows the network to track and transform temporal sequences of input signals.  In some cases, one neuronal layer with recurrent connectivity might in turn connect to another neuronal layer with recurrent connectivity; for example, Kenyon cells of the mushroom body receive input from olfactory projection neurons of the antennal lobe, which are connected to each other by inhibitory and excitatory interneurons \cite{Shang:2007cg,Olsen:2008kv}. Such cases can be analogised to hierarchically connected reservoirs. In biological systems, it is thought that transforming inputs into a higher-dimensional neural code in the `reservoir' increases the associative memory capacity of the network \cite{Marr1969theory}.
Moreover, it is known that for the efficient processing of information unfolding in time, which requires networks to dynamically keep track of past stimuli, the brain can implement ladders of neural populations with hierarchically organised `temporal receptive fields' \cite{yeshurun2017amplification}.

The same principles of dimensional expansion in space and/or time apply to artificial RC networks, depending on the non-linear transformation of the inputs into a representation useful for learning the task at the single linear output layer.
We focus here on a popular form of RC called Echo State Networks \cite{Jaeger2001echo}, where the reservoir is implemented as a RNN with a fixed, random synaptic connection matrix. This connection matrix is set so the input `echoes' within the network with decaying amplitude.  The performance of an Echo State Network depends on certain network hyper-parameters that need to be optimised through grid search or explicit gradient descent. Given that the dependence of the network's performance on such hyper-parameters is both non-linear and task-dependent, such optimisation can be tedious.

%More recently, reservoir computing (RC) has been employed as an alternative approach of simulating RNNs. In RC, a fixed `reservoir' is employed to transform a temporal input signal in such a way that only single layer output perceptron can be trained to solve a learning task. The advantage of RC is that the reservoir is a fixed system that is responding to input and can be either computationally or physically defined. Since it is fixed the reservoir is not trained and BPTT is not necessary making them much simpler to train. Critical to such an approach is how well the reservoir can non-linearly transform into a representation that can be easily solved with a linear output layer. A popular form of RC are Echo State Networks proposed by Jaeger\cite{Jaeger2001echo}, where the reservoir is represented by a RNN with a fixed, random synaptic connection matrix. This connection matrix is scaled such that its spectral radius is $\leq 1$ such that it exhibits 'echo state' property. This is where the reservoir state is excited by the input the network maintains a decaying 'echo' of this input. This internal memory and the inherent timescales of the network, which are related to the spectral radius and leaking rate of the hidden neurons, can be useful for a range of applications. 

Previous works have studied the dependence of the reservoir properties on the structure of the random connectivity adopted, studying the dependence of the reservoir performance on the parameters defining the random connectivity distribution, and formulating alternatives to the typical Erdos-Renyi graph structure of the network \cite{deng2007collective, rodan2010minimum, bacciu2018concentric}. In this sense, in \cite{rodan2010minimum} a model with a regular graph structure has been proposed, where the nodes are connected forming a circular path with constant shortest path lengths equal to the size of the network, introducing long temporal memory capacity by construction. \new{The memory capacity has been studied previously for network parameters such as the spectral radius ($\rho$) and sparsity; in general memory capacity is higher for $\rho$ close to 1 and low sparsity, but high memory capacity does not guarantee high prediction \cite{FARKAS2016MC, Marzen2017Memory}. ESNs are known to perform optimally when at the ``edge of criticality''\cite{Livi2018Edge}, where low prediction error and high memory can be achieved through network tuning.} 

More recently, models composed of multiple reservoirs have gathered the attention of the community. From the two ESNs with lateral inhibition proposed in \cite{xue2007decoupled}, to the hierarchical structure of reservoirs first analysed by Jaeger in \cite{jaeger2007discovering}, these complex architectures of multiple, multilayered reservoirs have shown improved generalisation abilities over a variety of tasks \cite{gallicchio2018deep, jaeger2007discovering, malik2016multilayered}.  In particular, the works \cite{gallicchio2017echo} \cite{gallicchio2018design} have studied different dynamical properties of such hierarchical structures of ESNs, \new{while \cite{MA202020DeepR} have proposed hierarchical (or deep) ESNs with projection encoders between layers to enhance the connectivity of the ESN layers}. \new{The partitioning (or modularity) of ESNs was studied by \cite{Rodriguez2019Modularity}, where the ratio of external to internal connections was varied. By tuning this partitioning performance can be increased on memory or recall tasks.} Here we demonstrate that one of the main reasons to adopt a network composed by multiple, pipelined sub-networks, is the ability to introduce multiple timescales in the network's dynamics, which can be important in finding optimal solutions for complex tasks. Examples of tasks that require such properties are in the fields of speech, natural language processing, and reward driven learning in partially observable Markov decision processes \cite{szita2006reinforcement}. A hierarchical structure of temporal kernels \cite{hermans2012recurrent}, as multiple connected ESNs, can discover higher level features of the input temporal dynamics. Furthermore, while a single ESN can be tuned to incorporate a distribution of timescales with a prefixed mode, optimising the system hyper-parameters to cover a wide range of timescales can be problematic.

Here, we show that optimisation of hyper-parameters can be guided by analysing how these hyper-parameters are related to the timescales of the network, and by optimising them according to the temporal dynamics of the input signal and the memory required to solve the considered task. This analysis improves performance and reduces the search space required in hyper-parameter optimisation.
In particular, we consider the case where an ESN is split into two sections with different hyper-parameters resulting in separate temporal properties. In the following, we will first provide a survey of timescales in ESNs before presenting the comparative success of these hierarchical ESNs on three different tasks. The first is the non-linear auto-regressive moving average 10 (NARMA10) task which requires both memory and fast non-linear transformation of the input. Second, we explore the performance of the network in a reconstruction and state ``perception'' task with different levels of external white noise applied on the input signal. Finally, we apply the hierarchical ESN to a permuted sequential MNIST classification task, where the usual MNIST hand written digit database is serialised and permuted as a 1d time-series. 

\section{Survey of timescales in Echo State networks}
We begin by describing the operations of an ESN and present a didactic survey of the inherent timescales in ESNs, which will be drawn upon in later sections to analyse the results.

As introduced in the previous section, an ESN is a recurrent neural network and 
the activity, $\mathbf{x}(t)$, of the neurons due to a temporal input signal $\mathbf{s}(t)$ is given by

%\begin{align}
%\textbf{x}(t+\delta t) & = (1-\alpha)\textbf{x}(t) \nonumber \\
%& +\alpha f\Big\{\gamma W_{\rm in}\textbf{s}+\dfrac{\rho}{|\lambda^{\mathrm{max}}_W|} W\textbf{x}(t)\Big\}
%\label{x}
%\end{align}

\begin{align}
 \mathbf{x}(t+\delta t) & = (1-\alpha)\mathbf{x}(t) +\alpha f\left( \mathbf{h}(t) \right),  \label{x} \\
 \mathbf{h}(t) & = \gamma \mathbf{W}_{\rm in}\mathbf{s}(t)+\rho \mathbf{W}\mathbf{x}(t), \label{eq:input} 
\end{align}

where $\mathbf{W}$ is a possibly sparse random matrix defining the connectivity of the network, $\mathbf{W}_{\rm in}$ defines the input adjacency matrix, and $\gamma$ is a rescaling factor of the input weights.  $\alpha=\delta t / \tau$ is the leakage term of the node, and $\rho$ is a scaling factor for the spectral radius of the connectivity matrix and will be discussed in more detail in the following. $f()$ is a non-linear function, which in this work we define as the hyperbolic tangent.
To ensure that the network exhibits the Echo-State property, and so that the activity does not saturate, the initial random connectivity matrix, $\mathbf{W}$, is rescaled by its maximum eigenvalue magnitude (spectral radius),  $|\lambda^{\mathrm{max}}_\mathbf{W}| = \max|\mathrm{eig}(\mathbf{W})|$, thus ensuring a unitary spectral radius which can be tuned using $\rho$ as a hyper-parameter. \new{In practice, $\mathbf{W}$ is constructed from a matrix of Normally distributed random numbers and the sparseness is enforced by randomly setting to zero a fixed proportion of these elements. Typically 10 non-zero connections per node are retained in $\mathbf{W}$.}

The timescales of this dynamical system are closely linked to the specific structure of $\mathbf{W}$ and to the two hyper-parameters; $\alpha$ and $\rho$. Since $\alpha$ is the leakage rate, it directly controls the retention of information from previous time steps, while $\rho$ specifies the maximum absolute magnitude of the eigenvalues and as such tunes the decay time of internal activity of the network.
Thus, the basic hyper-parameters that need to be set are $\gamma$, $\alpha$ and $\rho$. Considering the nonlinear dependence of the network performance on these values and the task-dependent nature of an efficient parameterisation, this process can be challenging. Such hyper-parameters are commonly optimised through a grid search or through explicit gradient descent methods in online learning paradigms \cite{jaeger2007optimization}. However, the fine tuning procedure can be guided, and the searchable space reduced, using a simple analysis of the hyper-parameters' relation to the timescales of the network, the external signal's temporal dynamics, and the memory required to solve the considered task.

\begin{figure}[tbh]
\begin{center}
    \includegraphics[width=1\textwidth]{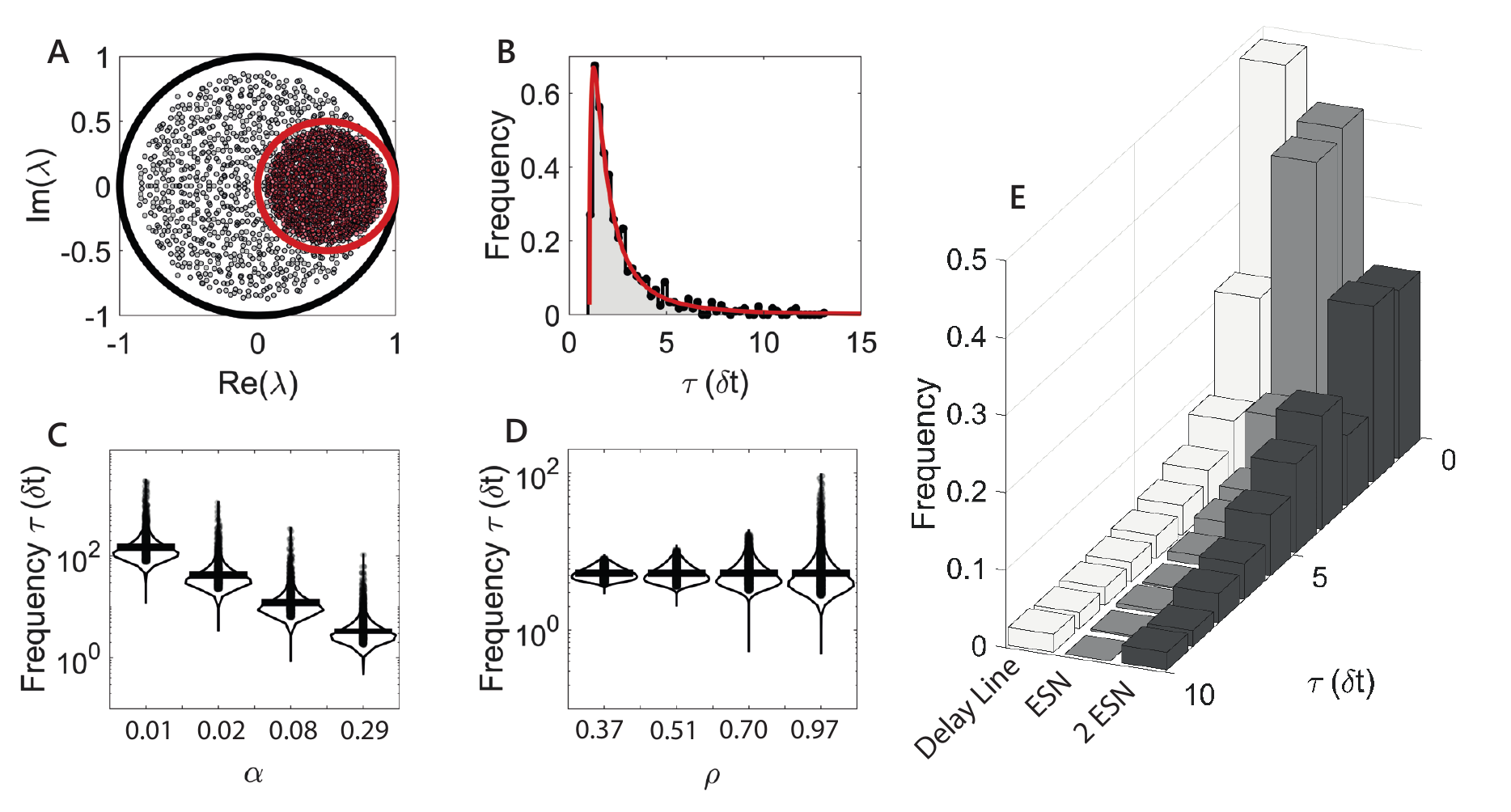}
    \caption{\small{The analysis of the timescales of the system in the linear regime can guide the search for the optimal values of the hyper-parameters $\alpha$ and $\rho$. \textbf{A}: Translation and scaling of the eigenvalues of the system due to the presence of the leakage factor. \textbf{B}: Example of distribution of timescales, computed analytically (red line) and computationally (black points) estimated from the eigenvalues of $\mathbf{W}$. \textbf{C}: Pirate plot of the distributions of timescales as $\alpha$ increases. Both axes are logarithmic. Higher $\alpha$ values correspond to longer timescales and to a more compressed range of timescales (logarithmic y-axis). \textbf{D}: Pirate plot of the distributions of timescales: as $\rho$ increases, the range of timescales expands. Again, both axes are logarithmic. \textbf{E}: Example distributions of timescales for reservoirs with different connectivity structure. From left to right, a delay line, single ESN, 2 ESNs (connected and unconnected, see text for the reason why the timescales for these two structures are the same in the linear regime). The higher complexity of the models reported is reflected in a richer distribution of timescales.}}
  \label{Figure1}
  \end{center}
\end{figure}

Considering that the eigenvalues $\lambda_{\mathbf{W}}$ of the connectivity matrix are inside the imaginary unit circle due to the normalisation procedure described previously, and that $\alpha$ is a constant common to all neurons, the eigenvalues of the linearised system given by Eq.\ \ref{x} are 

\begin{equation}
\lambda=1-\alpha(1-\rho\lambda_{\mathbf{W}}).
\label{lambda}
\end{equation}

This corresponds to a rescaling of value $\alpha\rho$ and to a translation of value $1-\alpha$ across the real axis of the original $\lambda_{\mathbf{W}}$. This operation on the eigenvalues of $\mathbf{W}$ is depicted in Fig.\ \ref{Figure1}A. Thus, considering that each eigenvalue $\lambda_i$ can be decomposed in its corresponding exponential decaying part $\exp(-\delta t/\tau_i)$ and its oscillatory imaginary component, the timescales of the linearised system are

\begin{align}
\tau &=\dfrac{\delta t}{1-\mathrm{Re}(\lambda)} \\
     &=\frac{\delta t}{\alpha(1-\rho \mathrm{Re}(\lambda_\mathbf{W}))}
\label{timescales}
\end{align}    

%\guido{then the matrix is NOT sparse, right? Why should we highlight so much the possibility of sparse connectivity if we do not make use of this?}
%Matt: In the simulations the it is sparse (on average 10 connections per node)

\new{When the connectivity matrix, $\mathbf{W}$, is given by a sparse matrix with non-zero elements drawn randomly from a uniform distribution with the range $[-1,1]$, then the corresponding eigenvalues will be uniformly distributed within a circle with a radius of $\max(|\lambda_{\mathbf{W}}|)$ in the complex plane \cite{girko1985circular}. These eigenvalues are then re-scaled by $\max(|\lambda_\mathbf{W}|)$ to ensure they are within the unit circle.} The distribution of the eigenvalues then reveals the distribution of timescales of the linearised system. Indeed, given $p\left(\mathrm{Re}(\lambda),\mathrm{Im}(\lambda)\right)$, the distribution of timescales can be found through computation of the marginal $p\big(\mathrm{Re}(\lambda))=\int p\big(\mathrm{Re}(\lambda),\mathrm{Im}(\lambda)\big) d\mathrm{Im}(\lambda)$ and the change of variable defined in equation \ref{timescales}\new{, giving 
\begin{equation}
p(\tau)=\dfrac{2 \delta t^2}{\pi \alpha^2 \rho^2\tau^2}\sqrt{\alpha^2\rho^2-\left(\alpha-\delta t/\tau\right)^2}
\label{eq:p_tau}
\end{equation}
}
Importantly we note that whilst the eigenvalues are uniformly distributed over the unit circle, the timescales are not due to the inverse relationship between them. The resulting distribution of the linearised system, shown in Fig.\ \ref{Figure1}B (red line), is in excellent agreement with the numerically computed distribution for a single ESN (black points + shaded area). 

The analytical form of the distribution, together with Eq.\ \ref{timescales}, allows us to explicitly derive how changes in $\alpha$ and $\rho$ affect the network timescales. Notably we can obtain analytical expression for the minimum, maximum and most probable (peak of the distribution) timescale: 

%Potentially we could add these as labels to Fig 1B

\begin{align}
    \tau_\text{min} & = \frac{\delta t}{\alpha(1+\rho)}, \label{eq:p_min} \\
    \tau_\text{max} & = \frac{\delta t}{\alpha(1-\rho)}, \label{eq:p_max} \\
    \tau_\text{peak} & = \frac{5\delta t}{4\alpha(1-\rho^2)} \left( 1 - \sqrt{1 - \frac{24}{25}(1-\rho^2) } \right) \label{eq:p_peak}
\end{align}
\new{where Eq.\ \ref{eq:p_max} and \ref{eq:p_min} can be derived directly from Eq.\ \ref{timescales}, while Eq.\ \ref{eq:p_peak} follows from maximisation of Eq.\ \ref{eq:p_tau}.}
As expected, $\alpha$ strongly affects all these three quantities; interestingly, though, $\alpha$ does not influence the relative range of the distribution, $\tau_\text{max} / \tau_\text{min} = (1+\rho)/(1-\rho)$. Indeed $\alpha$ plays the role of a unit of measure for the $\tau$s, and can then be used to scale the distribution in order to match the relevant timescales for the specific task. On the other hand, $\rho$ does not strongly affect the shape of the distribution, but determines how dispersed the $\tau$s are. Given the finite number of $\tau$s expressed by a finite ESN, the hyper-parameter $\rho$ can be used to balance the raw representation power of the network (how wide the range of timescales is) with the capacity to approximate any given timescale in that range. Fig.\ \ref{Figure1}C and D give a more detailed view of how the distribution of timescales changes as $\alpha$ and $\rho$, respectively, vary; note the logarithmic scale on the y-axis, that makes the dependence on $\alpha$ linear.  The link between the eigenvalues and the reservoir dynamics can be shown through the analysis of the network response to an impulsive signal, shown in Section \ref{Sec:Eig_th}, where the experimental activities are compared with the theoretical ones expected from the linearised system. 

\subsection{Hierarchical Echo-State Networks} \label{Sec.HESN}

\begin{figure}
    \centering
    \includegraphics[width=0.8\columnwidth]{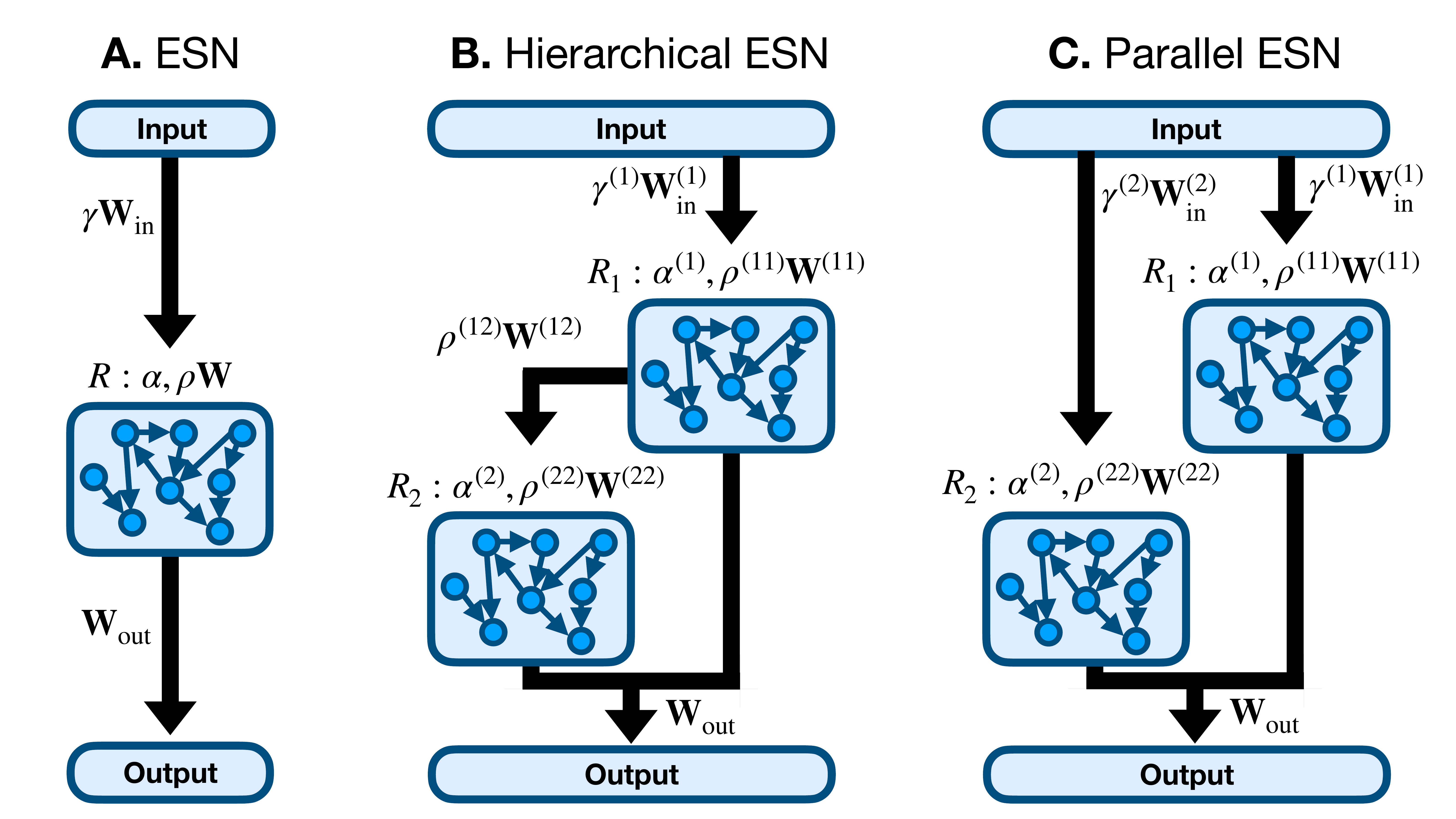}
    \caption{Single and hierarchical echo-state network (ESN) architectures. \textbf{A}: A single ESN with internally connected nodes with a single set of hyper-parameters $\alpha$ and $\rho$. \textbf{B}: A hierarchical ESN composed of 2 connected reservoirs where the input is fed into reservoir 1 only and the connection is unidirectional from R1 to R2\new{, which is identical to the 2 layer DeepESN of \cite{gallicchio2018design}}. \textbf{C}: A parallel (or unconnected hierarchical) ESN where the network is partitioned into 2 reservoirs, R1 and R2, which each receive the input and provide output but have distinct hyper-parameters. }
    \label{fig:multi_ESN}
\end{figure}

Different studies have proposed alternatives to the random structure of the connectivity matrix of ESNs, formulating models of reservoirs with regular graph structures. Examples include a delay line \cite{rodan2010minimum}, where each node receives and provides information only from the previous node and the following one respectively, and the concentric reservoir proposed in \cite{bacciu2018concentric}, where multiple delay lines are connected to form a concentric structure.  Furthermore, the idea of a hierarchical architecture of ESNs, where each ESN is connected to the preceding and following one, has attracted the reservoir computing community for its capability of discovering higher level features of the external signal \cite{Gallicchio2017DeepResComp}. Fig.\ \ref{fig:multi_ESN} schematically shows the architecture for (\textbf{A}) a single ESN, (\textbf{B}) 2 sub-reservoir hierarchical ESN for which the input is fed into only the first sub-reservoir which in turn feeds into the second and (\textbf{C}) a parallel ESN, where two unconnected sub-reservoirs receive the same input. \new{These heirarchical ESNs are identical to the 2 layer DeepESN given by \cite{gallicchio2018design}.} A general ensemble of interacting ESNs can be described by

\begin{align}
\mathbf{x}^{(k)}(t+\delta t) =  (1-\alpha^{(k)})\mathbf{x}^{(k)} 
+\alpha^{(k)} f\left( \mathbf{h}^{(k)}(t) \right), \label{eq:HESN_def}\\
\mathbf{h}^{(k)}(t) =  \gamma^{(k)} \mathbf{W}_{\rm in}^{(k)}\mathbf{s}^{(k)}(t) + \sum_{l}^{N_\text{ESN}} \rho^{(kl)} \mathbf{W}^{(kl)}\mathbf{x}^{(l)}(t), \label{eq:HESN_def2}
\end{align}

where the parameters have the similar definitions as in the case of a single ESN in Eq.\ \ref{x}. The index $k$ indicates the network number and $N_\text{ESN}$ is the total number of networks under consideration. In a hierarchical structure of ESNs $\mathbf{W}^{(kl)}\neq0$ for $k=l$ or $k=l+1$ only, and $\mathbf{W}^{(kl)}$ can be drawn from any desirable distribution thanks to the absence of feedback connections to higher-order reservoirs. Indeed, in this case, the necessary condition for the Echo-State network property is that all the inner connectivity matrices $\mathbf{W}^{(kk)}$ have eigenvalues with an absolute value less than one. Furthermore, in the typical hierarchical structure proposed in previous works \cite{jaeger2007discovering, gallicchio2018deep, gallicchio2018design, malik2016multilayered, sun2017deep}, the input is fed to the first network only, and $\mathbf{W}^{(k)}_{\rm in}\neq 0$ if $k=1$ only. We emphasise that the values of $\alpha^{(k)}$ and $\rho^{(k)}$, which are closely related to the timescales and repertoire of dynamics of network number $k$ (and, in the case of hierarchical reservoirs, also to all subsequent networks), do not have to be equal for each ESN, but can be chosen differently to fit the necessity of the task. In particular, some tasks could require memory over a wide range of timescales that could not effectively be covered by a single ESN.

In Fig.\ \ref{Figure1}E we show examples of the timescale distributions of the corresponding linearised dynamical systems for different ESN structures, from the simple delay line model to the higher complexity exhibited from two hierarchical ESNs. In order from left to right, the histograms of timescales are for a delay line, a single ESN, and two ESNs (whether hierarchically connected or unconnected; see below for clarification). All the models share an ESN with $\rho=0.9$ and $\alpha=0.9$; where present, the second reservoir has $\alpha=0.2$. By construction, the richness and range of timescales distributions reported increases with the complexity of the models. However, we note how a simple delay line could exhibit longer temporal scales than the other structures analysed thanks to its constant and high value of minimum path length between any pairs of nodes. Nevertheless, its limited dynamics restricts its application to simple tasks. The cases with two ESNs show a bimodal distribution corresponding to the two values of $\alpha$. 

Yet, the spectrum of the eigenvalues of the linearised system is only partially informative of the functioning and capabilities of an ESN. This is clearly demonstrated by the fact that a hierarchical and a parallel ESN share the same spectrum in the linear regime. Indeed, for a hierarchical ESN, whose connectivity matrix of the linearised dynamics is given by:
\begin{align}
    \tilde{\mathbf{W}}=\begin{bmatrix}
    \mathbf{W}^{(11)} & \mathbf{0} \\
    \mathbf{W}^{(21)} & \mathbf{W}^{(22)} 
  \end{bmatrix} ,
\end{align}
it is easy to demonstrate that every eigenvalue of $\mathbf{W}^{(11)}$ and $\mathbf{W}^{(22)}$ is also an eigenvalue of $\tilde{\mathbf{W}}$, irrespective of $\mathbf{W}^{(12)}$, not unlike what happens for a parallel ESN (where $\mathbf{W}^{(12)} = \mathbf{0}$, and hence the demonstration follows immediately). Nonetheless, as we will see in the next sections, the hierarchical ESN has better performance on different tasks compared to the other structures considered, including the parallel ESN.

%\st{The histograms of timescales shown for the double ESN structure is the same for the connected case and the unconnected reservoirs. Indeed, the eigenvalues of the linearised system describing the two reservoirs  }%The peak corresponding to higher $\delta t$ values in the distribution of timescales for the connected reservoirs is shifted in comparison to $ESN_2^{u}$. Of course, while the timescales shown for the unconnected case can be computed separately for the two reservoirs under consideration, the timescales of the connected structure are experimentally estimated through an analysis of the eigenvalues $\lambda_c$

% \begin{align}
   % \lambda_c=(\mathbb{I}_N-\tilde{\alpha})+\tilde{\alpha}\lambda_\tilde{W}, \\
    % \tilde{\alpha}=\begin{bmatrix}
    % \mathbb{I}_{N_1}\alpha^{(1)} & \textbf{0} \\
    % \textbf{0} & \mathbb{I}_{N_2}\alpha^{(2)}
   % \end{bmatrix}, \;
    % \tilde{W}=\begin{bmatrix}
    % W^{(11)} & \textbf{0} \\
    % W^{(21)} & W^{(22)}  
  % \end{bmatrix}
% \end{align}

%\st{would not change if $W^{(21)}$, describing the interaction between the two networks, is equal or not equal to zero. This consideration shows how the spectrum of the eigenvalues is only partially informative of the behaviour of the system, and demonstrates why the difference in performance between the connected and unconnected architectures will be increasing with the strength of interaction between the networks (See Section ref{Sec:narma}).}

It is interesting to note, in this respect, that the success of the hierarchical ESN is generally achieved when the leakage term of the first reservoir is higher than the leakage term of the second (or, in other words, when the first network has much shorter timescales). Such observation opens the way to an alternative route to understand the functioning of the hierarchical structure, as the first reservoir expanding the dimensionality of the input and then feeding the enriched signal into the second network. Indeed, in the following, we will show how, in a crude approximation and under the above condition of a wide separation of timescales, the first ESN extracts information on the short term behaviour of the input signal, notably its derivatives, and the second ESN integrates such information over longer times.

We begin with the (continuous time) linearized dynamics of a Hierarchical ESN is given by
\begin{align}
    \dot{\mathbf{x}}^{(1)}(t) &= -\mathbf{M}^{(1)} \mathbf{x}^{(1)}(t) +  \mathbf{W}_{\rm in}^{(1)} \mathbf{s}(t)  \label{eq.x1}, \\
    \dot{\mathbf{x}}^{(2)}(t) &= -\mathbf{M}^{(2)} \mathbf{x}^{(2)}(t) + \mathbf{W}^{(12)} \mathbf{x}^{(1)}(t),
\end{align}
where, for simplicity, we have reabsorbed the $\rho^{(kl)}$ and $\gamma^{(k)}$ factors into the definitions of $\mathbf{W}^{(kl)}$ and $\mathbf{W}_{\rm in}^{(k)}$ respectively, and the new constants can be derived with reference to Eq.~\ref{x} and \ref{eq:input}; for example:
\begin{equation}
    \mathbf{M}^{(k)} = \frac{\alpha^{(k)}}{\delta t} \, \big[1 - f'(0) \, \rho^{(k)} \, \mathbf{W}^{(kk)} \big].
\end{equation}
The neuron activity can be projected on to the left eigenvector of each of the $\mathbf{M}^{(i)}$ matrices. As such we define the eigenvector matrices, $\mathbf{V}^{(i)}$, where each row is a left eigenvector and so satisfies the equation $\mathbf{V}^{(i)} \mathbf{M}^{(i)} = \boldsymbol{\Lambda}^{(i)} \mathbf{V}^{(i)}$.  $\boldsymbol{\Lambda}^{(1)}$ and $\boldsymbol{\Lambda}^{(22)}$ are the diagonal matrices of the eigenvalues of the two $\mathbf{M}$ matrices. Using these we can define $\mathbf{y}^{(k)} \equiv \mathbf{V}^{(k)} \, \mathbf{x^{(k)}}$, and so the dynamical equations can be expressed as
\begin{align}
    \dot{\mathbf{y}}^{(1)}(t) &= -\boldsymbol{\Lambda}^{(1)} \, \mathbf{y}^{(1)}(t) +  \tilde{\mathbf{W}}_{\rm in}^{(1)} \mathbf{s}(t), \label{eq.y1} \\
    \dot{\mathbf{y}}^{(2)}(t) &= -\boldsymbol{\Lambda^}{(2)} \, \mathbf{y}^{(2)}(t) + \tilde{\mathbf{W}}^{(12)} \, \mathbf{y}^{(1)}(t), \label{eq.y2}
\end{align}
where $\tilde{\mathbf{W}}_{\rm in}^{(1)} = \mathbf{V}^{(1)} \mathbf{W}_{\rm in}^{(1)}$ and $\tilde{\mathbf{W}}^{(12)} = \mathbf{V}^{(2)} \, \mathbf{W}^{(12)} \, \left(\mathbf{V}^{(1)}\right)^{-1}$ are the input and connection matrices expanded in this basis. Taking the Fourier transform on both sides of Eq.~\ref{eq.y1}, such that $ FT\left[\mathbf{y}^{(1)}(t)\right] = \tilde{\mathbf{y}}^{(1)}(\omega)$ and $FT\left[\dot{\mathbf{y}}^{(1)}(t)\right] = - i \omega \tilde{\mathbf{y}}^{(1)}(\omega)$, where $i$ is the imaginary unit. The transform $\tilde{\mathbf{y}}^{(2)}(\omega)$ of $\mathbf{y}^{(2)}(t)$ can now be expressed as a function of the transform of the signal $\tilde{\mathbf{s}}(\omega)$ giving
\begin{align}
    (\boldsymbol{\Lambda}^{(1)} -i \omega \mathbf{I}) \tilde{\mathbf{y}}^{(1)}(\omega) &= \tilde{\mathbf{W}}_{\rm in}^{(1)} \tilde{\mathbf{s}}(\omega) \label{eq.fty1}
\end{align}
where $\mathbf{I}$ is the identity matrix of the same size as $\boldsymbol{\Lambda}^{(1)}$. 
If the second ESN's timescale are much longer than that of the first one (\textit{i.e.}, $\boldsymbol{\Lambda}^{(1)} \gg \boldsymbol{\Lambda}^{(2)}$), then we can expand the inverse of the $\tilde{\mathbf{y}}^{(1)}$ coefficient on the LHS of Eq.\ \ref{eq.fty1} when $\boldsymbol{\Lambda}^{(1)} \rightarrow \infty$ as
\begin{align}
  (\boldsymbol{\Lambda}^{(1)} -i \omega \mathbf{I})^{-1} & = ( \boldsymbol{\Lambda}^{(1)})^{-1} \left( 1 - i \omega (\boldsymbol{\Lambda}^{(1)})^{-1}\right)^{-1} \\
  & \approx (\boldsymbol{\Lambda}^{(1)})^{-1} \sum_{n=0}^\infty (i \omega (\boldsymbol{\Lambda}^{(1)})^{-1})^n
\end{align}
By applying this approximation to Eq.\ \ref{eq.fty1}, and by defining the diagonal matrix of characteristic times $\mathbf{T}^{(1)} \equiv -(\boldsymbol{\Lambda}^{(1)})^{-1}$, the relation between the activity of reservoir 1 and the input in Fourier space is given by
\begin{align}
  \tilde{\mathbf{y}}^{(1)}(\omega) &=  -\mathbf{T}^{(1)} \sum_{n=0}^\infty (-i \omega \mathbf{T}^{(1)})^n \tilde{\mathbf{W}}_{\rm in}^{(1)} \tilde{\mathbf{s}}(\omega).
\end{align}
The coefficients of this series are equivalent to taking successive time derivatives in Fourier space, such that $(-i\omega)^n \tilde{s} = d^{(n)} \tilde{s}/ dt^{(n)}$. So by taking the inverse Fourier transform  we find the following differential equation for $y^{(1)}$
\begin{align}
  \mathbf{y}^{(1)}(t) &=  -\mathbf{T}^{(1)} \sum_{n=0}^\infty (\mathbf{T}^{(1)})^n \tilde{\mathbf{W}}_{\rm in}^{(1)} \frac{d^{(n)} \mathbf{s}(t) }{ dt^{(n)}},
\end{align}
which can be inserted into Eq.\ \ref{eq.y2} to give
\begin{equation}
    \dot{\mathbf{y}}^{(2)} = \boldsymbol{\Lambda}^{(2)} \, \mathbf{y}^{(2)} - \tilde{\mathbf{W}}^{(12)} \, T^{(1)} \, \left[
     \tilde{\mathbf{W}}_{\rm in}^{(1)} \mathbf{s}(t) + \sum_{n=1}^\infty (\mathbf{T}^{(1)})^n \tilde{\mathbf{W}}_{\rm in}^{(1)} \frac{d^{(n)} \mathbf{s}(t) }{ dt^{(n)}} \right]. \label{Eq:HESN_rep}
\end{equation}
Thus the second ESN integrates the signal with a linear combination of its derivatives. In other words, the first reservoir expands the dimensionality of the signal to include information regarding the signal's derivatives (or, equivalently in discretised time, the previous values assumed by the signal). In this respect, Eq.\ \ref{Eq:HESN_rep} is key to understanding how the hierarchical connectivity between the two reservoirs enhances the representational capabilities of the system.
The finite-difference approximation of the time derivatives appearing in Eq.\ \ref{Eq:HESN_rep} implies that a combination of past values of the signal appears, going back in time as much as the retained derivative order dictates.

\subsection{Online learning of hyper-parameter}
Selecting the hyper-parameters of such systems can be challenging. Such selection process can be informed by the knowledge of the natural timescales of the task/signal at hand. Alternatively one can resort to a learning method to optimise the parameters directly. The inherent limitation of these methods is the same as learning the network weights with BPTT: the whole history of network activations is required at once. One way to by-pass this issue is to approximate the error signal by considering only past and same-time contributions, \new{as suggested by  Bellec {\it et al.} \cite{bellec2020solution} in their framework known as  {\it e-prop} (see also \cite{manneschi2020alternative}), and derive from this approximation an online learning rule for the ESN hyper-parameters. Following their approach, we end up with a novel learning rule for the leakage terms of connected ESNs that is similar to the rule proposed by Jaeger {\it et al.} \cite{jaeger2007optimization} but extended to two hierarchical reservoirs}. The main learning rule is given by:
\begin{equation}
    \frac{d E}{d \alpha^{(i)}}(t) = \sum_{k=1}^{N_\text{ESN}} \frac{\partial E}{\partial \mathbf{x}^{(k)}(t)} \mathbf{e}^{(ki)}(t) \label{Eq:Online}
\end{equation}
where $\mathbf{e}^{(ki)}(t) = d \mathbf{x}^{(k)}(t) / d \alpha^{(i)} $ is known as the eligibility trace \new{which tracks the gradient of neuron activities in the reservoir number $k$ with respect to the $i$-th leakage rate. Given the closed form for the hierarchical ESNs in Eqs.~\ref{eq:HESN_def} and \ref{eq:HESN_def2} these terms can be readily calculated. \new{For our $N_\text{ESN}$ sub-reservoirs in the hierarchical structure there will be $N_\text{ESN}^2$ eligibility traces to track how each sub-reservoir depends on the other leakage rates. In the hierarchical case of a fixed feed-forward structure some of these traces will be zero, and the number of non-zero eligibility traces would be $N(N+1)/2$}. Since the update of the neuron's activity depends on its previous values, so do the eligibility traces; therefore, they} can be calculated recursively through
\begin{align}
    \mathbf{e}^{(ki)}(t+\delta t)   = (1-\alpha^{(k)}) \mathbf{e}^{(ki)}(t)  
    & + \delta_{ki}( f(\mathbf{h}^{(k)}(t)) - \mathbf{x}^{(k)}(t) )  \nonumber \\
     & + \alpha^{(k)}f'(\mathbf{h}^{(k)}(t))\sum_{l\neq k} \rho^{(kl)} \mathbf{W}^{(kl)} \mathbf{e}^{(li)}(t) \label{eq:elig},
\end{align}
where $\delta_{ki} = 1$ if $ k=i$ and $0$ otherwise, i.e the Kronecker delta. The update of equations \ref{eq:elig} for each k-i pair needs to follow the order of dependencies given by the structure of connected reservoirs considered. The eligibility trace is an approximation that only includes same-time contributions to the gradient but has the advantage that is can be easily computed online. A complete description of our method is given in the Supplementary Material. For an example where the mean squared error function $E(t)=\frac{1}{2}\big[\tilde{y}(t)-y(t)\big]^2$ is used in a regression task and a structure composed by two reservoirs, the updating equations on the leakage terms are
\begin{align}
\alpha^{(1)}\leftarrow \alpha^{(1)}-\eta_{\alpha}\big[\tilde{y}(t)-y(t) \big] \mathbf{W}_{\rm out} \begin{pmatrix} \mathbf{e}^{(11)}(t) \\  \mathbf{e}^{(12)}(t)\end{pmatrix} \nonumber \\
 \alpha^{(2)}\leftarrow \alpha^{(2)}-\eta_{\alpha}\big[\tilde{y}(t)-y(t) \big] \mathbf{W}_{\rm out} \begin{pmatrix} \mathbf{e}^{(21)}(t) \\ \mathbf{e}^{(22)}(t) \end{pmatrix} \nonumber \\    
\end{align}
where $\eta_{\alpha}$ is the learning rate on the leakage terms and $\big(\mathbf{e}^{(k1)}(t),\mathbf{e}^{(k2)}(t)\big)$ ($k=1,2$ in this case with two reservoirs) is a vector composed by the juxtaposition of the eligibility traces, which can be computed through Eq.\ \ref{eq:elig}. Of course, the gradient can be combined with existing gradient learning techniques, among which we adopt the Adam optimiser, described in the Supplementary Material. \new{In all online learning simulations, training is accomplished through minibatches with updates at each time step. Training is stopped after convergence.} 
When learning $\alpha$s and the output weights simultaneously, the learning rates corresponding to these hyper-parameters need to be carefully set, since the weights need to adapt quickly to the changing dynamic of the network, but a fast convergence of $\mathbf{W}_{\rm out}$ can trap the optimisation process around sub-optimal values of the leakage terms. For a reservoir with trained and converged output weights, a further variation of $\alpha$'s, even in the right direction, could correspond to an undesirable increase in the error function. We found that this problem of local minimum can be avoided by applying a high momentum in the optimisation process of $\alpha$ and randomly re-initialising the output weights when the $\alpha$'s are close to convergence. The random re-initialisation functions to keep the output weights from being too close to convergence. Thus, we defined the convergence of the algorithm for $\alpha$'s as when the $\alpha$'s do not change considerably after re-initialisation. When this happens, it is possible to turn off the learning on the leakage terms and to optimise the read-out only.
More details about online training can be found in the discussions related to each task.  %The hyper-parameters used for the gradient descent learning are summarised in table \ref{tab:params}.

 %In the simulations when the output weights and the leakage terms are learnt simultaneously, we first trained $\alpha$s and $\mathbf{W}_{\rm out}$ until convergence of $\alpha$s, and then froze the values found for the leakage terms and learnt the output weights only. 

\section{Results}

The following sections are dedicated to the study of the role of timescales and the particular choices of $\alpha$ and $\rho$ in various tasks, with attention on networks composed by a single ESN, 2 unconnected ESNs and 2 hierarchical ESNs. The number of trainable parameters in each task for the different models will be preserved by using the same total number of neurons in each model. The results analysed will be consequently interpreted through the analysis of timescales of the linearised systems.

\vline
\subsection{NARMA10} \label{Sec:narma}

A common test signal for reservoir computing systems is the non-linear auto-regressive moving average sequence computed with a 10 step time delay (NARMA10) \cite{Atiya2000,Goudarzi2014comparative}. Here we adopt a discrete time formalism where $n = t/\delta t$ and the internal state of the reservoir is denoted as $\mathbf{x}_n = \mathbf{x}(n \delta t)$. The input, $s_n$, is a uniformly distributed random number in the range $[0,0.5]$ and the output time-series is computed using

\begin{equation}
  y_{n} =  y_{n-1} \left( a + b \sum_{k=1}^{D} y_{n-k} \right) + c s_{n-1} s_{n-D} + d,
\label{eq:narma}
\end{equation} 
where $D=10$ is the memory length, $a = 0.3$, $b = 0.05$, $c = 1.5$, and $d = 0.1$. The task for the network is to predict the NARMA10 output $y_n$ given the input $s_n$. We have adapted this to also generate a NARMA5 task where $D=5$ but the other parameters are unchanged. This provides an almost identical task but with different timescales for comparison.

\begin{figure}[tbh!]
 \begin{center}
    \includegraphics[width=1\columnwidth]{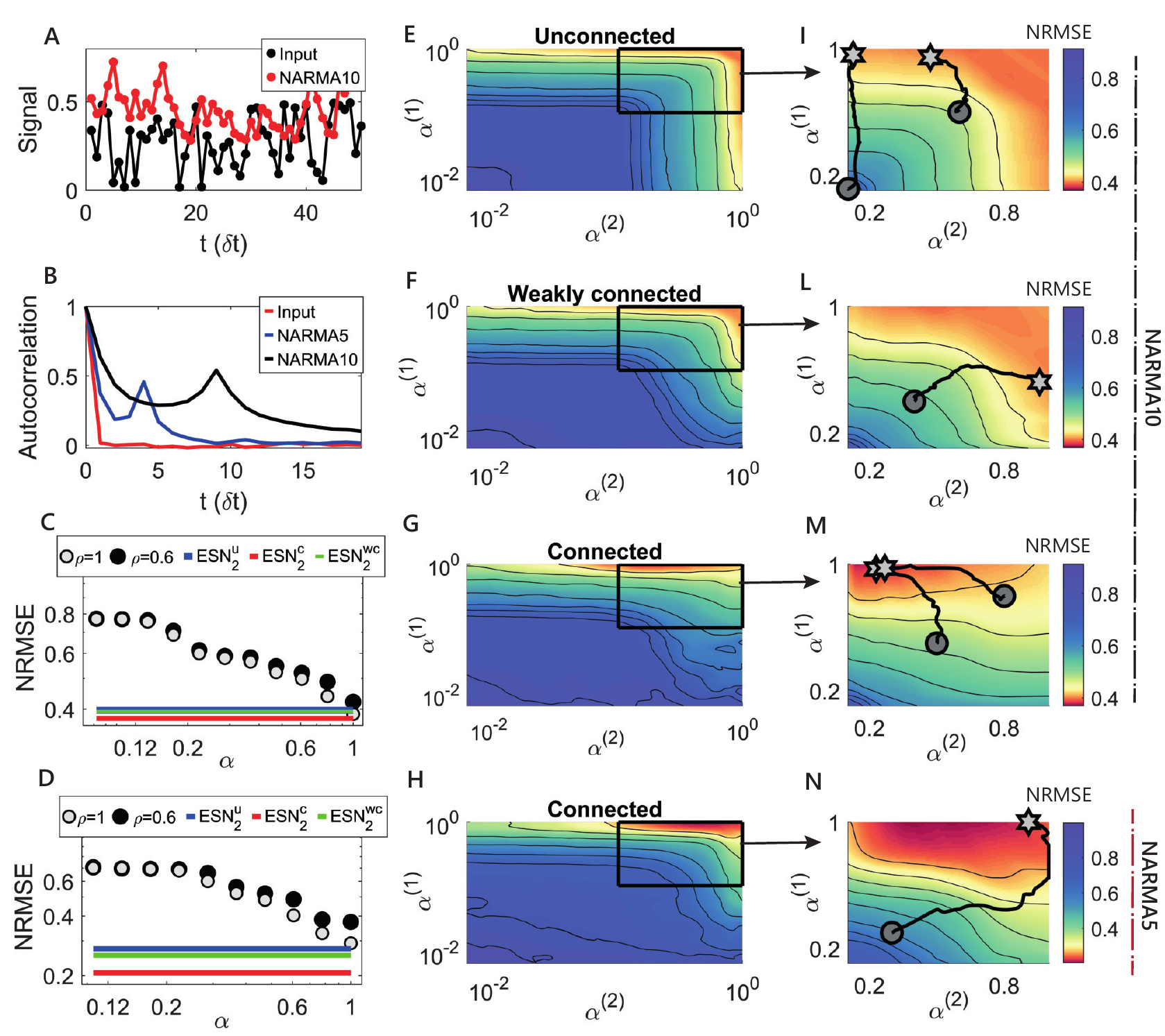}
    \caption{\small{Performance of single or hierarchical ESNs on the NARMA10 and NARMA5 task. \textbf{A}: Example input signal (black) and desired output (red) for the NARMA10 task. \textbf{B}: The auto-correlation function of the (black) input, (red) NARMA10 and (blue) NARMA5 desired output signals, showing a second peak at about 9 delay steps for the NARMA10 and 4 for the NARMA5.
    \textbf{C}: The NRMSE for a single ESN for with $\rho = 1.0$ and 0.63 over a range of $\alpha$. The NRMSE is lower for $\rho \approx 1$ and $\alpha = 1$. The solid lines show the minimum NRMSE for the unconnected (blue line) and connected (red line); for the unconnected case the minimum NRMSE is similar to the single ESN while the connected case has a smaller NRMSE by about 10\%.
    \textbf{D}: Average NRMSE of a single ESN for various $\alpha$ compared to the hierarchical ESNs for the NARMA5 task.}
    \textbf{E-N:} The average NRMSE surface using a hierarchical ESN computed for varying the leakage rates $\alpha^{(k)}$ of both the reservoir components for \textbf{E} and \textbf{I} (no coupling, \new{$\rho^{(12)}=0$}), \textbf{F} and \textbf{L} (weak coupling, \new{$\rho^{(12)}=0.1$}), and \textbf{G} and \textbf{M} (strong coupling, \new{$\rho^{(12)}=1$}). Panels \textbf{I-N} show a close up in region for the range $\alpha^{(k)} = [0.1,1]$ to highlight the changing behaviours. The lines on these panels show the trajectory of the $\alpha^{(k)}$ values trained directly using the online method. For each case of the coupling the online learning trends towards the approximate error minimum.
    \textbf{H} shows the NRMSE surface for the NARMA5 task using a strongly connected hierarchical ESN, with \textbf{N} again showing a zoom of the $\alpha=[0.1,1]$ region. The region of best performance is with $\alpha^{(2)}\approx0.5$ which matches the shorter timescale demonstrated in the auto-correlation in \textbf{B}. 
    %For the unconnected reservoir the NRMSE decreases symmetrically with $\alpha^{(1)}$ and $\alpha^{(2)}$, reporting a minimum at $\alpha^{(1)} = \alpha^{(2)} = 1$. For the weakly and strongly connected cases a lower NRMSE can be achieved with a lower value of $\alpha^{(2)}$, thus slower timescales, when the reservoirs are strongly and unidirectionally interacting. The best configuration corresponds to leakage terms that reflects the nature of the signal and the auto-correlation function of the desired output value, since a value of $\alpha^{(1)}\approx 1$ is necessary to follow the rapid changes of the process and a value of $\alpha^{(2)}\approx 0.1$ corresponds approximately to the second peak of the auto-correlation structure (red, panel \textbf{B}).
}
  \label{fig:narma}
   \end{center}
\end{figure}

The task of reconstructing the output of the NARMA10 sequence can be challenging for a reservoir as it requires both a memory (and average) over the previous 10 steps and fast variation with the current input values to produce the desired output. A typical input and output signal is shown in Fig.\ \ref{fig:narma}A and the corresponding auto-correlation function of the input and output in B. Since the input is a random sequence it does not exhibit any interesting features but for the output the auto-correlation shows a clear peak at a delay of 9 $\delta t$ in accordance with the governing equation. For a reservoir to handle this task well it is necessary to include not only highly non-linear dynamics on a short timescale but also slower dynamics to handle the memory aspect of the task. 

This regression task is solved by training a set of linear output weights to minimise the mean squared error (MSE) of the network output and true output. The predicted output is computed using linear output weights on the \new{concatenated} network activity ($\mathbf{x}_n = \left( \mathbf{x}^{(1)}_n, \mathbf{x}^{(2)}_n \right)^T$), such that
\begin{equation}
    \tilde{y}_n = \mathbf{x}_n^{T} \mathbf{W}_\mathrm{out}
\end{equation}
where $\mathbf{W}$ is the weight vector of length N+1 when an additional bias unit is included. The MSE is minimised by using the ridge regression method \cite{LUKOSEVICIUS2009} such that the weights are computed using
\begin{equation}
    \mathbf{W}_\mathrm{out} = \left( \mathbf{x}^T\mathbf{x} - \lambda \mathbf{I} \right)^{-1} \mathbf{x}^T \mathbf{y}
\end{equation}
where $\mathbf{x}$ is a matrix formed from the activation of the internal states with a shape of number of samples by number of neurons, $\mathbf{y}$ is the desired output vector, $\lambda$ is the regularisation parameter \new{that is selected using a validation data set} and $\mathbf{I}$ the identity matrix. 
To analyse the performance of the ESNs on the NARMA10 task we use the normalised root mean squared error as
\begin{equation}
    \mathrm{NRMSE} = \sqrt{\frac{1}{N_s}\frac{ \sum_n^{N_s} (\tilde{y}_n - y_n)^2}{\mathrm{Var}(\mathbf{y})}},
\end{equation}
where $\tilde{y}_n$ is the predicted output of the network and $y_n$ is the true output as defined by Eq.\ \ref{eq:narma}.

To test the effectiveness of including multiple time-scales in ESNs, we simulate first a single ESN with $N=100$ neurons and vary both $\alpha$ and $\rho$ to alter the time-scale distribution. Secondly, we simulate a hierarchical ESN split into 2 reservoirs each with $N=50$ neurons, where we vary $\alpha^{(1)}$ and $\alpha^{(2)}$ with $\rho^{(1)} = \rho^{(2)} = 0.95$. The input factor was set as $\gamma^{(1)} = 0.2$ and $\gamma^{(2)} = 0$ for the connected hierarchical ESN but when they are unconnected the input is fed into both, such that $\gamma^{(1)} = \gamma^{(2)} = 0.2$. In all cases the NRMSE is \new{computed on an unseen test set and} averaged over 20 initialisations of the ESN with a running median convolution is applied to the error surfaces to reduce outliers. In parallel to this we have also applied the online training method for the $\alpha$ hyper-parameters. The hyper-parameters used for the gradient descent learning are summarised in Table \ref{tab:params}.

Figure \ref{fig:narma}E-G and I-M show the NRMSE depending on $\alpha^{(1)}$ and $\alpha^{(2)}$ for 3 variations of the hierarchical ESN connection strength on the NARMA10 task. In the unconnected case ($\rho^{(21)} = 0$, panels E and I), we find that the NRMSE drops by increasing both leakage rates but the minimum is when one of the leakage rates is $\approx 0.5$. This is in agreement with the online learning method for the $\alpha$s in shown in I but the error minimum is shallow and prone to noise in the signal or ESN structure.  For the weakly connected hierarchical ESN (\new{$\rho^{(21)}=0.1$}, panels F and L) we find again that when the sub-reservoirs have different timescales the NRMSE is reduced. In comparison to the unconnected case the error surface is asymmetric with a minimum at approximately $\alpha^{(1)} = 1.0$ and $\alpha^{(2)} \approx 0.5$. As the strength of the connection is increased (\new{$\rho^{(21)}=1.0$}, Panel G and M), the minimum error moves to a lower leakage rate in the second reservoir ($\alpha^{(2)} \approx 0.2 $) which reflects a better separation of the timescale distributions. \new{This is a gradual effect with respect to the connection strength since stronger connection allows for a relative increase of the expanded input from the first reservoir compared to the base input signal.} Since the input feeds into reservoir 1, a high $\alpha$ provides a transformation on the input over short time-scales, expanding the dimensionality of the signal, offering a representation that preserves much of the dynamic of the driving input and that is fed to the second reservoir. Then, since the latter does not have a direct connection to the input it performs a longer timescale transformation of the internal states of reservoir 1. In this way the reservoirs naturally act on different parts of the task, i.e. reservoir 1 provides a fast non-linear transformation of the input while reservoir 2 follows the slower varying 10-step average of the signal, and thus returning a lower NRMSE.  \new{As a side note, we can demonstrate the validity of the theoretical analysis in Section \ref{Sec.HESN} by replacing the first reservoir by Eq.\ \ref{Eq:HESN_rep} on the NARMA task (see Section 3 Supplementary Material), resulting in a similar landscape as in Fig.\ \ref{fig:narma}G and a similar optimal value for} $\alpha^{(2)}$.

Figure \ref{fig:narma}C shows the relative performance of the single ESN to the minimum values for the unconnected ($\mathrm{ESN}^u_2$) and connected ($\mathrm{ESN}^c_2$) hierarchical reservoirs. The single ESN shows the similar decrease in NRMSE with increasing $\alpha$ and reaches a similar minimum NRMSE as the unconnected case. In comparison with the connected cases the multiple timescales provides a more optimised result. If we consider the analysis of the timescales discussed in the previous section the choice of these hyper-parameters becomes more evident. With $\alpha=1$ the timescale distribution of the network is sharply peaked close to the minimum timescale of 1 discrete step while when $\alpha=0.1$ this peak is broader and the peak of the distribution is closer to the second peak present in the auto-correlation function shown in Panel B. We note that whilst the most likely timescale is $\tau_\text{peak} \approx 6$ for $\alpha=0.1, \rho=0.95$ which is lower than the natural timescale of the problem, the increased width of the distribution increases the number of timescales at $\tau=10$ dramatically which maybe why a lower $\alpha$ is not necessary.

To further investigate the effect of the inherent timescale of the task on the timescales we performed a similar analysis on the NARMA5 task. Figure \ref{fig:narma}H and N show the NRMSE surface for the strongly connected case. The minimum error occurs at $\alpha^{(1)} \approx 1.0$ (similar to the NARMA10 results in G and M) but $\alpha^{(2)} \approx 0.5$ (as opposed to $\approx 0.2$ for NARMA10). This is due to the shorter timescales required by the NARMA5 task and the peak timescale for these values is much closer to the peak in the auto-correlation shown in B. Panel D shows the performance of the single ESN where again the optimal leakage rate is $\alpha=1$ and similar to the unconnected cases but the NRMSE is higher than the connected cases.

In this theoretical task where the desired output is designed {\it a priori}, the memory required and the consequent range of timescales necessary to solve the task are known. Consequently, considering the mathematical analysis in section \ref{Sec.HESN}, and that for hierarchical ESNs the timescales of the first ESN should be faster than those of the second Fig.\ \ref{fig:narma}), the best-performing values of the leakage terms can be set {\it a priori} without the computationally expensive grid search reported in Fig.\ \ref{fig:narma}E-I. However, it can be difficult to guess the leakage terms in the more complex cases where the autocorrelation structure of the signal is only partially informative of the timescales required. 

This problem can be solved using the online learning approach defined through Eq.\ \ref{Eq:Online}. \new{In this case, learning is accomplished through minibatches and the error function can be written explicitly as
\begin{align}
E(t)=\dfrac{1}{2\rm N_{batch}} \sum_{m=1}^{\rm N_{batch}} \big[\tilde{y}(t,m)-y(t,m)\big]^2 \label{eq:MSE_online}   
\end{align}
where $\rm N_{batch}$ is the minibatch size and m is its corresponding index. A minibatch is introduced artificially by dividing the input sequence into $\rm N_{batch}$ signals or by generating different NARMA signals. Of course, the two methods lead to equivalent results if we assure that the $\rm N_{batch}$ sequences are temporally long enough.  }
A learning rate $\eta_{\alpha}/\eta_{W} \approx 10^{-2}-10^{-3} $ was adopted. The optimiser used for this purpose is Adam, with the suggested value of $\beta_1=0.9$ adopted for the output weights and a higher first momentum $\beta_1=0.99$ adopted for the leakage terms. Instead, we set $\beta_2=0.999$ of the second momentum for both types of parameters (See section \ref{Sec:online_training} for a description of the updating rules).  
Panels I-N show a zoomed in region of the error surface with the lines showing the online training trajectory of the $\alpha$ hyper-parameters. In each case the trajectory is moving towards the minimum NRMSE of the $\alpha$ phase space.

\setlength\extrarowheight{4pt}
\sisetup{range-phrase=-}
\begin{center}
\begin{table}[tbh!]
\begin{tabular}{ |p{0.22\linewidth}|p{0.2\linewidth}|p{0.2\linewidth}|  }
\hline
\multicolumn{3}{|c|}{Learning hyper-parameters} \\
\hline
\hline
%\multicolumn{2}{|c|}{} &  \multicolumn{2}{c|}{} \\
%\multicolumn{2}{|c|}{\textbf{NARMA}} &  \multicolumn{2}{c|}{\textbf{psMNIST}} \\
 %\multicolumn{2}{|c|}{} &  \multicolumn{2}{c|}{} \\
            & \textbf{NARMA/Telegraph}       & \textbf{psMNIST} \\
 \hline
 Network size, $\rm N$                &  $100$               & $1200$   \\
 Minibatch size, \new{$\rm N_{batch}$}          &  $10$                &  $50$     \\
 \hline
 & \multicolumn{2}{c|}{Learning $\mathbf{W}_{\rm out}$}  \\
 \hline
$\eta_W $           &  $10^{-3}$   &  $ \SI{5e-2}{} \left[\SI{5e-3}{}\right]^\dagger$    \\
$\beta_1$           &  $0.9$               &  $0.9$   \\
$\beta_2$           &  $0.999$             & $0.999$   \\
 $\epsilon$         &  $10^{-8}$           &  $10^{-8}$\\
 \hline
    &  \multicolumn{2}{c|}{Learning $\alpha$} \\
 \hline
$\eta_{\alpha}$     &  $5\times 10^{-6}$   &  $10^{-4}$     \\
$\beta_1$           &  $0.99$              &  $0.999$   \\
$\beta_2$           &  $0.999$             & $0.999$   \\
 $\epsilon$         &  $10^{-8}$           &  $10^{-8}$\\
 \hline
\end{tabular}
\caption{\small{Table of the hyper-parameters adopted in the online learning process. $\eta$ is the learning rate in each case, while $\beta_1, \beta_2$ and $\epsilon$ are parameters for the Adam optimiser (further details are given in the Supplementary Material). The $^\dagger$ symbol indicates that the learning rate $\SI{5e-2}{}$ is for the case with $4$ hidden states, while the learning rate $\SI{5e-3}{}$ is for the case with $28$ hidden states. This decrease of $\eta$ is due to the increase in the dimensionality of the representation for the latter case in comparison to the situation where the read-out is composed by four concatenated values of activity. Furthermore, such learning rates are $10$ times higher than the case in which only the read-out is trained (only in the psMNIST task). Thus, the high learning rate adopted has the purpose to introduce noise in the learning process and to avoid local minima in the complex case where $\alpha$ and $\mathbf{W}_{\rm out}$ are optimised simultaneously.  }} \label{tab:params}
\end{table}
\end{center}

\subsection{A volatile environment} \label{Sec:th_task}

We now turn to study the reservoir performance on a task of a telegraph process in a simulated noisy environment. The telegraph process $s^{(1)}(t)$ has two states that we will call {\it up} (1) and {\it down} (0), where the probability of going from a {\it down} state to an {\it up} state  $p(s=1|s=0)$ (or the opposite $p(s=0|s=1)$) is fixed for any time step. The environment is also characterised by a telegraph process $s^{(2)}(t)$, but the transition probability is much lower and controls the transition probability of the first signal. 
To simplify the notation in the following we denote the probability of the signal $i$ transitioning from state $a$ to state $b$ as $P(s^{(i)}(t)=a|s^{(i)}(t-\delta t)=b)=p^{(i)}_{ab}(t)$.
The signal taken under consideration is then composed by a fast telegraph process with probabilities $p^{(1)}_{01}(t)$ and $p^{(1)}_{10}(t)$, whose values are interchanged by following the dynamic of a slower telegraph process $s^{(2)}(t)$. Every time the slower environment signal changes its state, the probabilities of the first signal are changed, i.e  $p^{(1)}_{01}(t) \leftrightarrow p^{(1)}_{10}(t)$. The resulting signal is then characterised by

\begin{align}
p^{(1)}_{10}(t) & = \begin{cases} p_1, & \mbox{if } s^{(2)}(t)=0 \\ p_2, & \mbox{if } s^{(2)}(t)=1 \end{cases} \label{Eq:Tel1} \\
p^{(1)}_{01}(t) & = \begin{cases} p_2, & \mbox{if } s^{(2)}(t)=0 \\ p_1, & \mbox{if } s^{(2)}(t)=1 \label{Eq:Tel2} \end{cases}
\end{align}

The transition probabilities of the second signal are fixed and symmetric such that
\begin{equation}
    p^{(2)}_{01}(t) = p^{(2)}_{10}(t) = p_3 \label{Eq:Tel3},
\end{equation}

The probabilities $p_1, p_2$ and $p_3$ are fixed parameters of the signal that define the process. Given that the second signal controls the probabilities of the first telegraph process, we say that it defines the regime of the input, while we refer to the {\it up} and {\it down} values of the first process simply as states. Thus, the reconstruction of $s^{(1)}(t)$ \new{from the input will be called state reconstruction, while reconstruction of $s^{(2)}(t)$ will be called regime reconstruction. These reconstructions can be considered separately or as a joint task requiring the system to be modeled on long and short timescales simultaneously.} Due to the probability transition caused by $s^{(2)}(t)$, both states and regime will be equally present over a infinitely long signal. The values adopted for the simulation are $p_1=0.05$, $p_2=0.1$ and  $p_3=0.0005$. %Whilst the transition rate for the second signal is symmetric, thus over a infinitely long signal both states will be equally present, the first signal is not symmetric, i.e $p_1 \neq p_2$. Therefore the value of the second signal will cause a measurable difference in the first signal. \matt{We should detail the probabilities used for generating the task}

\begin{figure}[h!]
 \centering
    \makebox[\textwidth][c]{\includegraphics[width=1\textwidth]{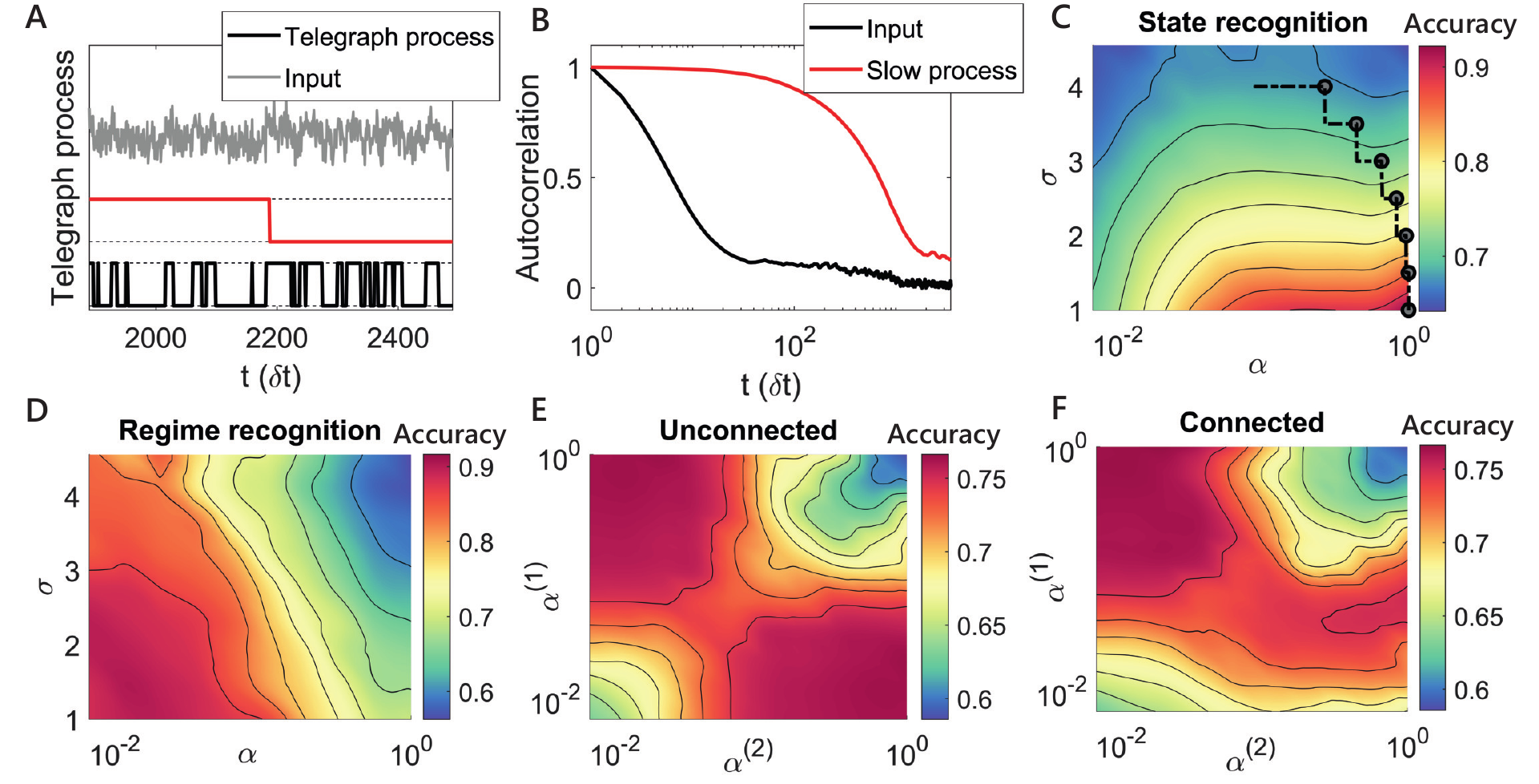}}
    \caption{\small{The best structure and parameters of the model depend on the specific environment considered, that is different values of the additive noise in the input signal, and on the specific desired output. \textbf{A}: Example of input signal and of its generative processes, which have a faster and a slower dynamic respectively. When the slower process \new{(red line)} is {\it up} ({\it down}), the other signal is in a regime where the average time in the zero (one) state is greater than the average time spent in the other state. The input signal (grey line) corresponds to the faster process (black line) with additional white noise. \textbf{B}: Auto-correlation structure of the two generative processes. 
    \textbf{C}: The accuracy surface for a single ESN on the state recognition sub-task for varying level of noise ($\sigma$) and leakage rate of the network showing that for increasing levels of noise a lower leakage rate is needed to determine the state. The line shows the trajectory of $\alpha$ using the online learning method when the strength of the noise is changed.
    \textbf{D}: The accuracy for a single ESN on the regime recognition sub-task for varying noise and leakage rate. In this case the low leakage rate is preferred for all values of noise.
    \textbf{E}: Accuracy surface for the state recognition sub-task for an \new{unconnected} hierarchical ESN showing how either of the leakage rates must be low while the other is high.
    \textbf{F}: Accuracy surface for the regime recognition sub-task for a hierarchical ESN showing the first reservoir must have a high leakage rate and the second a low leakage rate.
    %\textbf{C}, \textbf{F} and \textbf{I}: the online learning process when the signal probabilities are increase or decreased periodically. }}
    %\textbf{D,G} Performance of the model when it needs to reconstruct the faster telegraph process (\textbf{C}) or the slower one (\textbf{D}). In the \textbf{C} panel, it is clear how the best leakage term depends on the level of additional white noise ($\sigma$) applied; while an input with low $\sigma$ needs only to copied by the dynamic of the network, a noisier signal needs to be smoothed by integrators with slower timescales in order to efficiently solve the task. The \textbf{D} panel shows how the demand to reconstruct a different and slower process is reflected in the lower timescales that must be adopted by the reservoir in comparison to the task in \textbf{C}. \textbf{E-F} Performance for the task where the two processes must be reconstructed at the same time for connected ESNs (\textbf{E}) and unconnected ESNs (\textbf{F}).  
    }}
  \label{fig:Tel_part1}
\end{figure}

The input signal corresponds to $s^{(1)}(t)+\sigma \mathcal{N}(0,1)$, that is the faster telegraph process with additional white noise. The input signal constructed is a metaphor of a highly stochastic environment with two states and two possible regimes that define the probability of switching between the two states. The reservoir will be asked to understand in which state ($s^{(1)}(t)=1$ or 0) and/or regime ($s^{(2)}(t)=1$ or 0) it is for each time $t$, measuring the understanding of the model to estimate the state of the input signal. The input signal and telegraph processes is shown in Fig.\ \ref{fig:Tel_part1}A, while the B shows the corresponding auto-correlation structure of the processes. The auto-correlation shows that the input has a temporal structure of around 10 $\delta t$ while the slow `environment' process has a structure close to 1000 $\delta t$. This corresponds directly to the timescales defined by the probabilities of the signals. 

\begin{figure}[h!]
 \centering
    \makebox[\textwidth][c]{\includegraphics[width=1\textwidth]{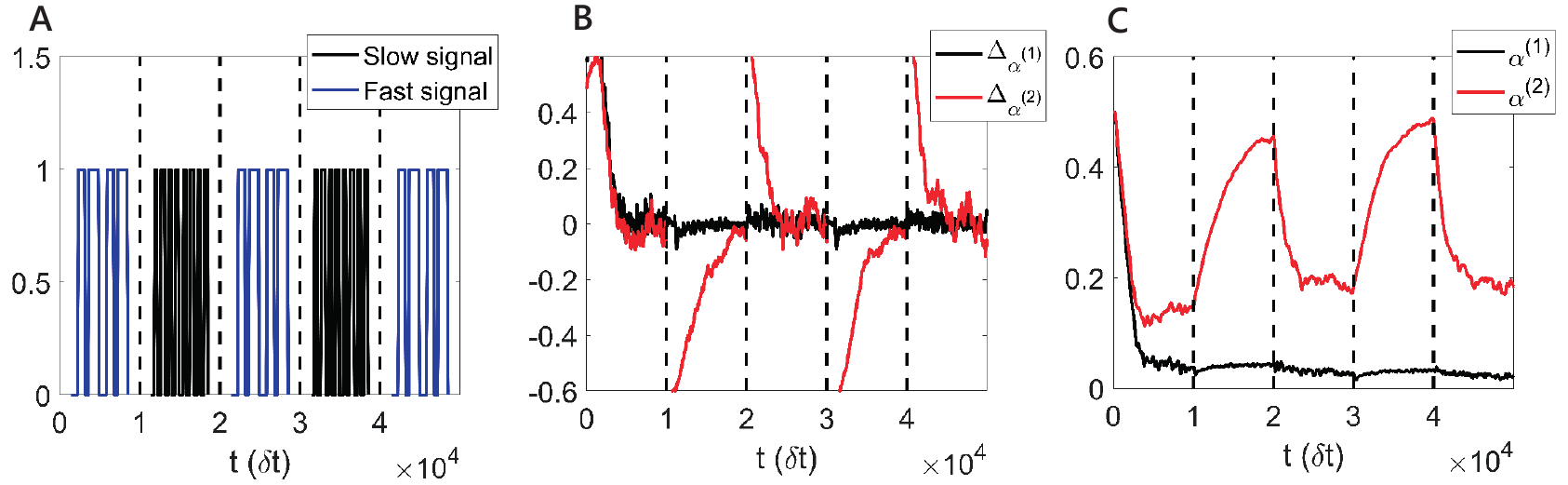}}
    \caption{\small{ The online training of the leakage terms can adapt to the changing environment, that is the signal probabilities are increased or decreased periodically.
    \textbf{A}: Scheme of the change of the values of probabilities, where high probabilities of switching are referred to as fast phase of the telegraph process, while low probabilities as slow phase.
    \textbf{B}: Running average of the gradients of $\alpha^{(1)}$ and $\alpha^{(2)}$ as time varies.
    \textbf{C}: Online adaptation of the leakage terms.
    }}
  \label{fig:Tel_part2}
\end{figure}

Panels C and D of Fig.\ \ref{fig:Tel_part1} show the performance of a single ESN when it is tasked to reconstruct the processes $s^{(1)}(t)$ \new{(state recognition)} and $s^{(2)}(t)$ \new{(regime recognition)} respectively. \new{In this simulation, learning is always accomplished online and the error function is the same as Eq. \ref{eq:MSE_online}. } First, panel C demonstrates how the leakage term, $\alpha$, must be tuned to the level of noise of the environment, and how lower values of $\alpha$ are desirable for noisier signals, in order to solve the state recognition problem. Indeed, the need to smooth the fluctuations of the input signal increases with $\sigma$, while for low values of noise the network should simply mimic the driving input. Second, panel D shows how the desirable values of $\alpha$ must be lower in the case where the network is asked to reproduce the slower dynamic of $s^{(2)}(t)$ independently of having to output the fast signal, in order to solve the regime recognition problem. \new{ This result exemplifies how the timescales of the network must be tuned depending on the desired output. It demonstrates that, even in this relatively simple environment, it is crucial to adopt multiple timescales in the network to obtain results that are robust with respect to a variation of the additional white noise $\sigma$.} 

Finally, panels E and F of Fig.\ \ref{fig:Tel_part1} show the accuracy of two unconnected (E) and connected (F) reservoirs when the network has to classify the state and the regime of the input signal at the same time. In this case, the desired output corresponds to a four dimensional signal that encodes all the possible combinations of states and regimes; for instance, when the signal is in the state one and in the regime one, we would require the first dimension of the output to be equal to one and all other dimensions to be equal to zero, and so on. The best performance occurs when one leakage term is high and the other one is low and in the range of significant delays of the auto-correlation function. \new{ This corresponds to one network solving the regime recognition and the other network solving the state recognition.} For the unconnected reservoirs, it does not matter which reservoir has high vs. low leakage terms, reflected by the symmetry of Fig.\ \ref{fig:Tel_part1}E, while for the connected reservoirs, the best performance occurs when the first reservoir has the high leakage term and the second the low leakage terms, see Fig.\ \ref{fig:Tel_part1}F, similar to Fig.\ \ref{fig:narma}. Both two-reservoir networks can achieve accuracy $0.75$, but the single ESN can not solve the task efficiently, \new{since it cannot simultaneously satisfy the need for high and low $\alpha$s}, reporting a maximum performance of about $0.64$.

%\new{This is due to the fact that the two reservoirs solve different parts of the problem. One reservoir solves the state recognition while the other solves the regime recognition. A high performance cannot be reached by a single reservoir: Fig.~\ref{fig:Tel_part1} C and D demonstrates that different time scales are needed to optimally solve each task separately. In the case of the two unconnected reservoirs, either reservoir can solve either task, leaving the remaining task to the other, as reflected by the symmetry of Fig.~\ref{fig:Tel_part1} E. In the case of the two connected reservoirs (Fig.~\ref{fig:Tel_part1} F) symmetry breaks and the first reservoir takes a high $alpha$ value (state recognition) with the second acquiring the lower $alpha$ value (regime recognition).} 
%high accuracy is achieved by one sub-reservoir solving the state recognition part of the task while the other solves the regime recognition part. The accuracy is not solely achieved by one sub-reservoir having a high $\alpha$ since this would be poor on the regime recognition part (as shown in Fig.\ref{fig:Tel_part1}D). Since both reservoirs receive the same input either sub-reservoir can have a high $\alpha$, but the other must have a lower value. When they are connected, only reservoir 1 receives the input, which breaks the symmetry of either sub-reservoir having a high $\alpha$, and so the best accuracy is achieved when sub-reservoir 1 has a high $alpha$.}

The path reported in panel C of Fig.\ \ref{fig:Tel_part1} and all panels in Fig.\ \ref{fig:Tel_part2} show the application of the online training algorithm in this environment. The values of the hyper-parameters adopted in the optimisation process through the Adam optimiser are the same as in section \ref{Sec:narma}, where we used a slower learning rate and a higher first momentum on the leakage terms in comparison to the values adopted for the output weights. The line of panel C (Fig.\ \ref{fig:Tel_part1}) shows the online adaptation of $\alpha$ for a simulation where the external noise increases from one to four with six constant steps of 0.5 equally spaced across the computational time of the simulation. The result shows how the timescales of the network decrease for each increase in $\sigma$, depicted with a circle along the black line. The path of online adaptation reports a decrease of the $\alpha$ value for noisier external signals. This result occurs because as the signal becomes noisier ($\sigma$ rises), it becomes more important to dampen signal fluctuations. This result also shows that the online algorithm can adapt in environments with varying signal to noise ratio. Panels A, B, C of Fig.\ \ref{fig:Tel_part2} show the online training of $\alpha^{(1)}$ and $\alpha^{(2)}$ for an environment composed by a faster and a slower composition of telegraph processes. This specific simulation is characterised by the alternation of two signals defined by Eq.\ \ref{Eq:Tel1}, \ref{Eq:Tel2} and \ref{Eq:Tel3}, each with different values of $p_1$ and $p_2$. In particular, while $p_1=0.5$ and $p_2=0.1$ for the 'fast' phase of the external signal, $p_1=0.1$ and $p_2=0.05$ for the 'slow' phase. In contrast, the slower timescale of the task defined by $p_3=0.0005$ remains invariant across the experiment. Panel C shows the adaptation of the leakage terms for this task in the case of a hierarchical structure of ESNs. While $\alpha^{(2)}$ adapts to the change of $p_1$ and $p_2$ following the transition between the two phases of the external signals, the relatively constant value of $\alpha^{(1)}$ indicates how the first network sets its timescales to follow the slower dynamic of the signal, characterised by the constant value of $p_3$. Thus, the composed network exploits the two reservoirs separately, and the first (second) reservoir is used to represent the information necessary to recognise the regime (state) of the external signal. \\ %\matt{Are you sure that $\alpha1$ and $\alpha2$ are used the right way around here? Would a very low $\alpha1$ smooth out the input for R2} \hl{AL: I agree with Matt, I didn't follow the reasoning here}    

\subsection{Permuted Sequential MNIST} \label{Sec:psMNIST}

The Permuted Sequential MNIST (psMNIST) task is considered a standard benchmark for studying the ability of recurrent neural networks to understand long temporal dependencies. The task is based on the MNIST dataset, which is composed of $60,000$ handwritten digits digitised to 28x28 pixel images. In the standard MNIST protocol every pixel is presented at the same temporal step so a machine has all the information of the image available at once and needs to classify the input into one out of ten classes. In contrast, in the psMNIST task, the model receives each pixel sequentially once at a time, so that the length of the one dimensional input sequence is $784$. Thus, the machine has to rely on its intrinsic temporal dynamic and consequent memory ability to classify the image correctly. Furthermore, each image in the dataset is transformed through a random permutation of its pixels in order to include temporal dependencies over a wide range of input timescales and to destroy the original images' structure. Of course, the same permutation is applied on the entire dataset. The performance of ESNs on the MNIST dataset, where each columns of pixels in a image is fed to the network sequentially (each image corresponds to a $28$ dimensional signal of length $28$ time steps), has been analysed in \cite{schaetti2016echo} and in \cite{manneschi2019sparce}. In \cite{schaetti2016echo} the original dataset was preprocessed through reshaping and rotating the original image to enhance the network's ability to understand high level features of the data. In this case, the original dataset is used.  In \cite{manneschi2019sparce}, the addition of thresholds and the introduction of sparse representation in the read-out of the reservoir was used to improve the performance of the network in the online learning of the standard MNIST task through reservoir computing.  
This section is focused on the analysis of the performance of ESNs on the psMNIST task and on their dependence on the range of timescales available in the network, i.e. the values of $\alpha$ and $\rho$ chosen. In contrast to the previous sections where ESNs are trained through ridge regression, we have applied an online gradient descent optimisation method. The cost function chosen to be minimised is the cross entropy loss

\begin{equation}
E=-\dfrac{1}{\rm N_{batch}}\sum_{m=1}^{\rm N_{batch}} \sum_{j=1}^{\rm N_{class}} \Big[ y_{j}(m) \log\big(\tilde{y}_{j}(m) \big)+\big(1-y_j(m)\big)\log\big(1-\tilde{y}_j(m)\big) \Big],
\end{equation}
\new{where m is the minibatch index, $\rm N_{batch}$ corresponds to the minibatch size and $\rm N_{class}$ is the number of classes.}
For this task the desired output, $y_j$, is a one-hot encoded vector of the correct classification while the desired output is a sigmoid function of the readout of the reservoir nodes. Furthermore, instead of reading out the activity of the reservoir from the final temporal step of each sequence only, we have expanded the reservoir representation by using previous temporal activities of the network. In practice, given the sequence of activities $\mathbf{x}(0), \mathbf{x}(\delta t),...,\mathbf{x}(\delta t T)$ ($T=784$) that defines the whole temporal dynamic of the network subjected to an example input sequence, we trained the network by reading out from the expanded vector $\mathbf{X}=\big[\mathbf{x}(M\delta t),\mathbf{x}(2M\delta t),...,\mathbf{x}(T\delta t)\big]$, where $M$ defines the 'time frame' used to sample the activities of the evolution of the system across time.
\begin{equation}
    \tilde{\mathbf{y}} = \text{sigm}\left(\sum_{n=1}^{T/M} \mathbf{W}_\text{out}^{(n)} \mathbf{x}(nM \delta t ) \right)
\end{equation},
where sigm stands for sigmoid activation function.
We then repeat the simulation for two different time frames of sampling for each different model, that is a single ESN and a pair of unconnected or connected ESNs, as in the previous sections.

The two values of $M$ used are $28$ and $196$, corresponding to a sampling of $28$ and $4$ previous representations of the network respectively. Of course, a higher value of $M$ corresponds to a more challenging task, since the network has to exploit more its dynamic to infer temporal dependencies. We note, however, that none of the representation expansions used can guarantee a good understanding of the temporal dependencies of the task, or in other words, can guarantee that the system would be able to discover higher order features of the image, considering that these features depend on events that could be distant in time. 

\begin{figure}[h!]
 \centering
    \includegraphics[width=1\textwidth]{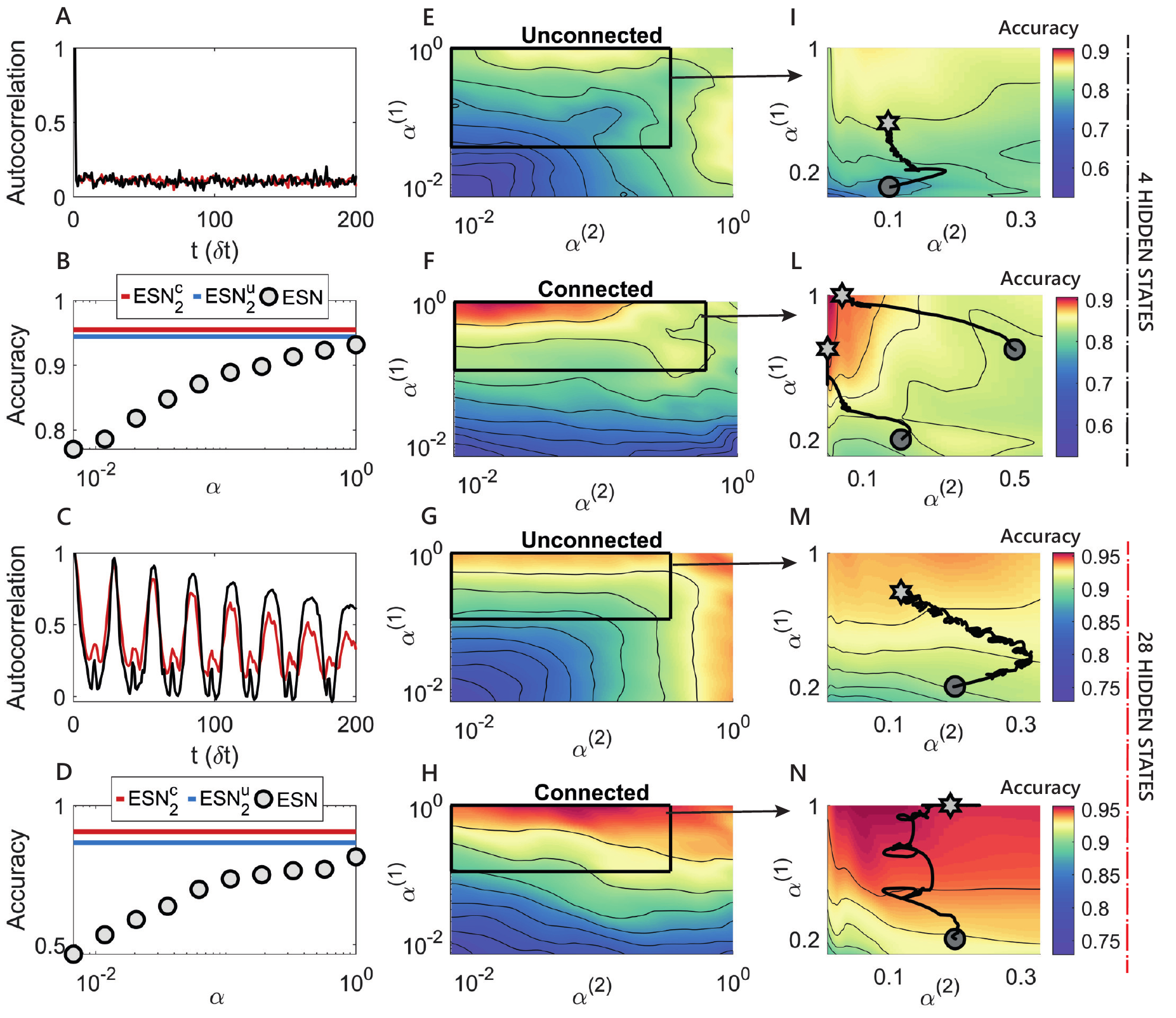}
    \caption{\small{The additional non linearity added by the hierarchical reservoir structure is responsible for a relevant modification and increase of the performance surface. \textbf{A,C}: Auto-correlation structure of the MNIST dataset for two examples of digits, where each pixel is presented one after the other (\textbf{C}), and auto-correlation structure of the data after the random permutation(\textbf{A}). The oscillatory trend in \textbf{C} reflects the form of the written digits, when this is seen one pixel after the other. The auto-correlation function of the permuted data is low, but not negligible, for all the temporal steps, showing the necessity to have a wide repertoire of timescales in the interval corresponding to the image size. \textbf{B,D}: Accuracy of a single ESN for various $\alpha$ values compared the maximum accuracy of the hierarchical ESNs with 4 hidden states (\textbf{B}) or 28 hidden states (\textbf{D)}. \textbf{E-F}: Case with low sampling frequency of the ESNs which corresponds to a higher demand of internal memory in the reservoir. While the best region of accuracy for the unconnected reservoirs is characterised by intermediate values of the leakage factors, the hierarchically connected network structure reports the best performance when the second network has slower dynamics. \textbf{G-H}: The utilisation of a high sampling frequency alleviates the need for long term memory, and the reservoirs prefer the regions with fast timescales. In both cases analysed, the additional complexity of the hierarchical model leads to a considerable boost in performance. \textbf{I-N}: Paths (black line, starting from the circle and ending in the star) that describe the online changes of the leakage terms achieved through the online training algorithm in a zoomed region of the performance surface of $\alpha^{(1)}$ and $\alpha^{(2)}$. The paths are smoothed through a running average. }}
  \label{fig:psmnist}
\end{figure}

In Fig.\ \ref{fig:psmnist} we again analyse the performance of two connected or unconnected ESNs varying $\alpha^{(1)}$ and $\alpha^{(2)}$ for both $M=28$ and $196$. In contrast to the previous sections, we now \new{use gradient descent learning on the output weights instead of ridge regression and } increase the total number of neurons in each model to $N=1200$ due to the complexity of the task. \new{The Adam optimiser is used; its parameters, for both the output weights and $\alpha$ learning, are in Table \ref{tab:params}. As previously, we have trained the output weights over a range of fixed $\alpha$s and report the performance on an unseen test data set. In parallel to this we have trained both the output weights and $\alpha$ values which, as shown by the lines on the contour plots, converge towards the minimum computed using the fixed $\alpha$'s.}

As in the other simulations, we found that the values of $\rho$ corresponding to the best performance was approximately one, which maximises the range of timescales and the memory available in the network. Fig.\ \ref{fig:psmnist}E-F shows the case with $M=28$, while Fig.\ \ref{fig:psmnist}G-H reports the accuracy for the simulation with $M=196$ where E and G are unconnected and F and H connected reservoirs. The accuracy surface demonstrates how, in the case of the unconnected ESNs with a fast sampling rate in panel G, the best performance is achieved when at least one of the two values of $\alpha$ is close to one. The result is due to the fast changing dynamic of the temporal sequence that is introduced through the random permutation of the pixels. On the contrary, in the case of the unconnected ESNs with a slow sampling rate in panel E the best accuracy is in a range of intermediate timescales since both partitions must respond to both fast and slow timescales. 

This relatively simple behaviour of the dependence of the accuracy on the setting of the hyper-parameters changes in the cases of two connected ESNs, whose additional complexity corresponds to a considerable increase in the performance. Fig.\ \ref{fig:psmnist}H reports how the network prefers a regime with a fast timescale in the first reservoir and a intermediate timescale in the second, which acts as an additional non-linear temporal filter of the input provided by the first network. The need of memory of events distant in time is emphasised in \ref{fig:psmnist}F, where the best performing network is composed by reservoirs with fast and slow dynamics respectively. The performance boost from the panels E-G to the ones F-H has only two possible explanations: first, the timescales of the second network are increased naturally thanks to the input from the first reservoir; second, the connections between the two reservoirs provide an additional non-linear filter of the input that can be exploited to discover higher level features of the signal. Thus, we can conclude once again that achieving high performance in applying reservoir models requires (1) additional non-linearity introduced through the interconnections among the reservoirs and (2) an appropriate choice of timescales, reflecting the task requirements in terms of external signal and memory.

Panels I, L, M and N show the application of the online training of $\alpha$s for the various cases analysed. In the psMNIST task we found that the major difficulties in the application of an iterative learning rule on the leakage terms are: the possibility to get trapped in local minima, whose abundance can be caused by the intrinsic complexity of the task, the intrinsic noise of the dataset, the randomness of the reservoir and of the applied permutation; the high computational time of a simulation that exploits an iterative optimisation process on $\alpha s$ arising from a practical constraint in the implementation. Indeed, while the activities of the reservoir can be computed once across the whole dataset and then saved in the case of untrained values of $\alpha $s, the activities of the nodes need to be computed every time the leakage terms change in the online learning paradigm. However, we found that using a higher learning rate $\eta_W$ on the output weights, compared to the value adopted in the paradigm where the leakage terms are not optimised (as in Panels E, F, G and H), can introduce beneficial noise in the learning process and help to avoid local minima. Furthermore, a higher value of the learning rate on the output weights corresponds to an increased learning rate on the thresholds, as shown from Eq.\ \ref{Eq:leakage_learning} and from the dependence of the updating equations on $\textbf{W}_{\rm out}$. As in the previous simulations of Sections \ref{Sec:narma} and \ref{Sec:th_task}, the output weights are randomly reinitialised after the convergence of $\alpha$s, helping the algorithm to avoid an undesirable quick convergence of weights. The online process is then ended when the leakage terms remain approximately constant even after the re-initialisation. Following this computational recipe, it possible to avoid the difficulties found and train the leakage terms efficiently.

Finally, we note how the best accuracy of $0.96$ reached throughout all the experiments on the psMNIST is comparable to the results obtained by recurrent neural networks trained with BPTT, whose performance on this task are analysed in \cite{chandar2019towards} and can vary from $0.88$ to $0.95$. In comparison to recurrent structures trained through BPTT, a network with two interacting ESNs provide a cheap and easily trainable model. However, this comparison is limited by the necessity of recurrent neural networks to carry the information from the beginning to the end of the sequence, and to use the last temporal state only or to adopt attention mechanisms.     

%\begin{table}
%\centering
%\begin{tabular}{|l|l|l|l|l|l|l|}
%\hline
%        & \multicolumn{3}{l|}{Single ESN} & \multicolumn{3}{l|}{Multi ESN} \\ \hline
%Task    & $\alpha$       & $\rho$      & NRMSE      & $\alpha^{(1)}$      & $\alpha^{(2)}$      & NRMSE      \\ \hline
%1 &         &          &            &         &         &            \\ \hline
%2      &         &          &            &         &         &            \\ \hline
%3 &         &          &            &         &         &            \\ \hline
%\end{tabular}
%\caption{}

%\end{table}

 \section{Conclusion}
 
 In summary, ESNs are a powerful tool for processing temporal data, since they contain internal memory and time-scales that can be adjusted via network hyper-parameters. Here we have highlighted that multiple internal time-scales can be accessed by adopting a split network architecture with differing hyper-parameters. We have explored the performance of this architecture on three different tasks: NARMA10, a benchmark composed by a fast-slow telegraph process and PSMNIST. In each task, since multiple timescales are present the hierarchical ESN performs better than a single ESN when the two reservoirs have separate slow and fast timescales. We have demonstrated how choosing the optimal leakage terms of a reservoir can be aided by the theoretical analysis in the linear regime of the network, and by studying the auto-correlation structure of the input and/or desired output and the memory required to solve the task. The theoretical analysis developed needs to be considered as a guide for the tuning of the reservoir hyper-parameters, and in some specific applications it could be insufficient because of the lack of information about the nature of the task. In this regard,  we showed how to apply a data-driven online learning method to optimise the timescales of reservoirs with different structures, demonstrating its ability to find the operating regimes of the network that correspond to high performance and to the best, task-dependent, choice of timescales. 
 The necessity of adopting different leakage factors is emphasised in the case of interactive reservoirs, whose additional complexity leads to better performance in all cases analysed. Indeed, the second reservoir, which acts as an additional non linear filter with respect to the input, is the perfect candidate to discover higher temporal features of the signal, and it consequently prefers to adopt longer timescales in comparison to the first reservoir, which has instead the role of efficiently representing the input. \new{ We believe such hierarchical architectures  will be useful for addressing complex temporal problems and there is also potential to further optimise the connectivity between the component reservoirs by appropriate adaptation of the online learning framework presented here.} 
 
 \clearpage
  \section{Appendix} 
  
  \subsection{Online Learning} \label{Sec:online_training}
\begin{figure}[tbh]
\begin{center}
    \includegraphics[width=1.\textwidth]{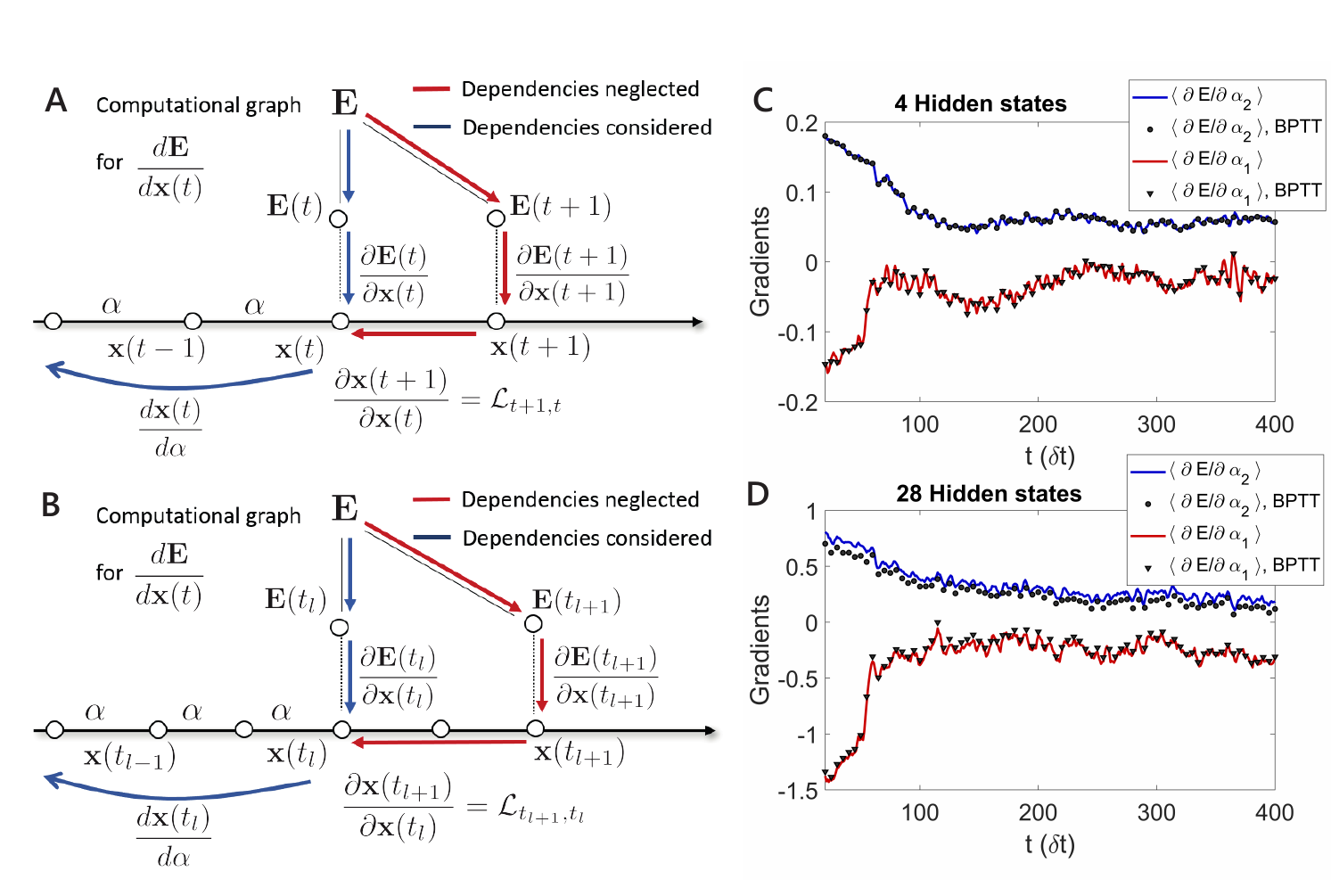}
    \caption{\small{The online training algorithm used \cite{bellec2020solution} maintains the majority of the temporal information of BPTT, while the approximation becomes less precise as the number of concatenated steps increases. \textbf{A-B}: Scheme of the computational graph for the contribution of $\frac{d E}{d \mathbf{x}(t)}$ for the case where the error function is 'continuous' across time \textbf{A}, and the case where the error function is sparse across time \textbf{B}. The blue arrows represent the factors considered, while the red arrows correspond to the factors that are neglected in the approximation. Each mathematical term adjacent to an arrow is a multiplicative factor in the contribution of a path of dependencies in the computation of $\frac{d E}{d \mathbf{x}(t)}$. \textbf{C-D}: Comparison of a running average of the derivative estimated through the online training algorithm used (red and blue lines) and BPTT (dots and triangles). The approximation is less precise when the number of hidden states used for the read-out increases, as it is evident from the greater distance between the blue trend and the dots in panel $\textbf{D}$.     }}
  \label{Fig:eprop_scheme}
  \end{center}
\end{figure}

  The online learning method formulated is similar to the approach followed in e-prop by \cite{bellec2020solution} (see also \cite{manneschi2020alternative}), a local learning rule for recurrent neural networks that exploits the concept of an eligibility trace, and in \cite{jaeger2007optimization}. As in these previous works, we approximated the error function to neglect the impact that the instantaneous and online changes of the network's parameters have on future errors.
 In particular, considering a recurrent neural network as the one depicted in the computational graph in Fig.\ \ref{Fig:eprop_scheme}A  and the dependencies of the error function $E$ on the activities $\mathbf{x}(t)$
 
 \begin{align}
    \dfrac{d E}{d \alpha} &=\sum_t \dfrac{d E}{d \mathbf{x}(t)}\dfrac{d \mathbf{x}(t)}{d \alpha} \nonumber \\ 
    &=\sum_t \Bigg\{\dfrac{\partial E(t)}{\partial \mathbf{x}(t)}+\dfrac{\partial E(t+1)}{\partial \mathbf{x}(t+1)}\dfrac{\partial \mathbf{x}(t+1)}{\partial \mathbf{x}(t)}+\dfrac{\partial E(t+2)}{\partial \mathbf{x}(t+2)}\dfrac{\partial \mathbf{x}(t+2)}{\partial \mathbf{x}(t+1)}\dfrac{\partial \mathbf{x}(t+1)}{\partial \mathbf{x}(t)}+... \Bigg\}\dfrac{d \mathbf{x}(t)}{d \alpha} \nonumber \\
    &=\sum_t \Bigg\{\sum_{t^{'}\geq t}\dfrac{\partial E(t^{'})}{\partial \mathbf{x}(t^{'})}\mathcal{J}_{t^{'}t} \Bigg\}\dfrac{d \mathbf{x}(t)}{d \alpha}, \label{Eq:bptt1} \\
    \mathcal{J}_{t^{'}t} & =\dfrac{\partial \mathbf{x}(t')}{\partial \mathbf{x}(t'-1)}\cdots\dfrac{\partial \mathbf{x}(t+1)}{\partial \mathbf{x}(t)} \label{Eq:bptt2}
 \end{align}
 
Eq. \ref{Eq:bptt1} and \ref{Eq:bptt2} define the algorithm back-propagation through time, where the dependencies of $\dfrac{d E}{d \mathbf{x}(t)}$ on activities at future time $t'$ do not permit the definition of an online learning rule. As in the works of \cite{jaeger2007optimization} and \cite{bellec2020solution} we decided to adopt the following approximation
 
\begin{align}
 \dfrac{d E}{d \alpha} &=\sum_t \Bigg\{\sum_{t'\geq t}\dfrac{\partial E(t^{'})}{\partial \mathbf{x}(t^{'})}\mathcal{J}_{t^{'}t} \Bigg\}\dfrac{d \mathbf{x}(t)}{d \alpha}  \nonumber \\
 &\approx \sum_t \dfrac{\partial E(t)}{\partial \mathbf{x}(t)}\dfrac{d \mathbf{x}(t)}{d \alpha} \label{Eq:eprop}
\end{align} 

We will now derive the equations defining the iterative learning approach for the example cost function 

 \begin{equation}
     E(t)=\frac{1}{2} \big[\tilde{y}(t)-y(t)\big]^2
 \end{equation}
 
 where $\tilde{y}$ is the desired output and $y=\mathbf{W}_{\rm out} \mathbf{x}(t)$ is the output of the ESN. Then, we desire to compute $\partial E/ \partial \alpha^{(k)}$, which describes the leakage term $k$ for a network compose by multiple reservoirs. 
 In particular, the case of two connected ESNs in considered and analysed here, while the more general case with N interacting ESNs can be easily derived following the same approach.
 In this case, the vector of activities $\mathbf{x}(t)=\big(\mathbf{x}_1(t),  \mathbf{x}_2(t) \big)$ is composed by the juxtaposition of the vectors of activities of the two reservoirs.
 \begin{align}
     \mathbf{x}^{(1)}(t+\delta t) & =(1-\alpha^{(1)})\mathbf{x}^{(1)}(t) + \alpha^{(1)}f\big[\mathbf{W}_{\rm in} \mathbf{s}(t)+\mathbf{W}^{(11)}\mathbf{x}^{(1)}(t)\big] \\
     \mathbf{x}^{(2)}(t+\delta t) &=(1-\alpha^{(2)})\mathbf{x}^{(2)}(t) + \alpha^{(2)}f\big[\mathbf{W}^{(21)} \mathbf{x}^{(1)}(t)+\mathbf{W}^{(22)}\mathbf{x}^{(2)}(t)\big]
 \end{align}
\begin{align}
 \dfrac{d E(t)}{d \alpha^{(1)}}=-\big[\tilde{y}(t)-y(t) \big] \mathbf{W}_{\rm out} \begin{pmatrix}\dfrac{d  \mathbf{x}^{(1)}(t)}{d \alpha^{(1)}} \\  \dfrac{d  \mathbf{x}^{(2)}(t)}{d \alpha^{(1)}}\end{pmatrix} \nonumber \\
 \dfrac{d E(t)}{d \alpha^{(2)}}=-\big[\tilde{y}(t)-y(t) \big] \mathbf{W}_{\rm out} \begin{pmatrix}\dfrac{d  \mathbf{x}^{(1)}(t)}{d \alpha^{(2)}} \\  \dfrac{d  \mathbf{x}^{(2)}(t)}{d \alpha^{(2)}}\end{pmatrix} \nonumber \label{Eq:leakage_learning}\\
\end{align}

 \begin{align}
     \dfrac{d  \mathbf{x}^{(1)}(t)}{d \alpha^{(1)}} & =(1-\alpha^{(1)})\dfrac{d  \mathbf{x}^{(1)}(t-1)}{d \alpha^{(1)}}-\mathbf{x}^{(1)}(t-1)+ \nonumber \\ 
     & +\alpha^{(1)}f'\big[\mathbf{W}_{\rm in} \mathbf{s}(t)+\mathbf{W}^{(11)} \mathbf{x}^{(1)}(t) \big] \mathbf{W}^{(11)}\dfrac{d  \mathbf{x}^{(1)}(t-1)}{d \alpha^{(1)}}+ \nonumber \\
     & +f\big[ \mathbf{W}_{\rm in} \mathbf{s}(t)+\mathbf{W}^{(11)} \mathbf{x}^{(1)}(t) \big] \\ \nonumber\\
     \dfrac{d  \mathbf{x}^{(2)}(t)}{d \alpha^{(2)}} & =(1-\alpha^{(2)})\dfrac{d  \mathbf{x}^{(2)}(t-1)}{d \alpha^{(2)}}-\mathbf{x}^{(2)}(t-1)+ \nonumber \\ 
     & +\alpha^{(2)}f'\big[\mathbf{W}^{(21)} \mathbf{x}^{(1)}(t)+\mathbf{W}^{(22)} \mathbf{x}^{(2)}(t) \big] \mathbf{W}^{(22)}\dfrac{d  \mathbf{x}^{(2)}(t-1)}{d \alpha^{(2)}}+ \nonumber \\
     & +f\big[ \mathbf{W}^{(21)} \mathbf{x}^{(1)}(t)+\mathbf{W}^{(22)} \mathbf{x}^{2)}(t) \big] \\ \nonumber\\
     \dfrac{d  \mathbf{x}^{(1)}(t)}{d \alpha^{(2)}} &=0 \\ \nonumber\\
     \dfrac{d  \mathbf{x}^{(2)}(t)}{d \alpha^{(1)}} & =(1-\alpha^{(2)})\dfrac{d  \mathbf{x}^{(2)}(t-1)}{d \alpha^{(1)}}+ \nonumber \\ 
     & +\alpha^{(2)}f'\big[\mathbf{W}^{(21)} \mathbf{x}^{(1)}(t)+\mathbf{W}^{(22)} \mathbf{x}^{(2)}(t) \big] \big[\mathbf{W}^{(22)}\dfrac{d \mathbf{x}^{(2)}(t-1)}{d \alpha^{(1)}}+ \nonumber \\ 
     & +\mathbf{W}^{(21)}\dfrac{d \mathbf{x}^{(1)}(t-1)}{d \alpha^{(1)}} \big]  
 \end{align}
 
 \new{That can be computed online tracking the eligibility traces $\dfrac{d \mathbf{x}^{(1)}(t)}{d \alpha^{(1)}}=\mathbf{e}^{(11)}(t)$, $\dfrac{d \mathbf{x}^{(2)}(t)}{d \alpha^{(1)}}=\mathbf{e}^{(21)}(t)$, $\dfrac{d \mathbf{x}^{(2)}(t)}{d \alpha^{(2)}}=\mathbf{e}^{(22)}(t)$ and updating them in an iterative way. Of course, for the more general case of $N$ connected reservoirs, the number of eligibility traces to be computed would be $N^2$. We note how the differences between the connected and unconnected reservoirs are: $\mathbf{e}^{(21)}(t)=0$ in the latter case, since the activity of the second reservoir does not depend on the activities of the first; $\mathbf{e}^{(22)}(t)$ would have an analogous expression to $\mathbf{e}^{(11)}(t)$ in the case of unconnected reservoirs.}
 
 In order to understand the meaning of the approximation in Eq.\ \ref{Eq:eprop}, we can consider the psMNIST task defined in section \ref{Sec:psMNIST}, in which two different numbers of previous hidden states are used for classification. In this example, the future terms $t'$ from which $\dfrac{d E}{d \mathbf{x}(t)}$ depends correspond to the concatenated temporal steps $\big\{t_l\big\}_{l=1,...,N_{conc}}$ used for the readout. Following the computational graph in panel B of Fig.\ \ref{Fig:eprop_scheme} , the approximation of BPTT is
 
 \begin{align}
 \dfrac{d E}{d \alpha} & =\sum_{l} \Bigg\{\sum_{q\geq l}\dfrac{\partial E(t_q)}{\partial \mathbf{x}(t_q)}\mathcal{J}_{t_q t_l} \Bigg\}\dfrac{d \mathbf{x}(t_l)}{d \alpha} \nonumber \\
 & \approx \sum_l \dfrac{\partial E(t_l)}{\partial \mathbf{x}(t_l)}\dfrac{d \mathbf{x}(t_l)}{d \alpha}
\end{align} 

where the contribution of the terms corresponding to $\sum_{q>l}\dfrac{\partial E(t_q)}{\partial \mathbf{x}(t_q)}\mathcal{J}_{t_q t_l}$ are neglected. The number of these terms increases as the number of hidden states used to define the read-out rises, and the contribution of the matrices $\mathcal{J}_{t_q t_l}$ becomes more important when the hidden states utilised are in closer proximity. Thus, the approximation used to define the online training algorithm is less precise for an increasing number of hidden states used. This consideration can be observed in Panels C and D of Fig.\ \ref{Fig:eprop_scheme}, in which the  values of the gradients are compared to those given by BPTT for the two different numbers of concatenated values adopted in Section \ref{Sec:psMNIST}.

 Given the gradients with respect of the parameters of the network $\frac{d E}{d \alpha^{(k)}}$ and $\frac{d E}{d W_{ij}}$ ($\mathbf{W}$ are the output weights here) in our simulations, we used the Adam optimisation algorithm, described below for completeness for a general parameter $\alpha$ (that could be one of the leakage terms or $W_{ij}$). 
 
 \begin{align}
    m_t &\leftarrow (1-\beta_1) m_{t-1}+\beta_1 \dfrac{d E}{d \alpha} \\
    v_t &\leftarrow (1-\beta_2) v_{t-1}+\beta_2 \big(\dfrac{d E}{d \alpha}\big)^2 \\
    \tilde{m}_t &\leftarrow m_t/(1-\beta_1^t) \\
    \tilde{v}_t &\leftarrow v_t/(1-\beta_2^t) \\
    \alpha_t&=\alpha_{t-1}-\eta_{\alpha} (\tilde{m}_t/(\sqrt{\tilde{v}_t}+\epsilon))
 \end{align}
 
 where $t$ is the index corresponding to the number of changes made and $m_0=0$, $v_0=0$.

 \subsection{Timescales, oscillations and eigenvalues} \label{Sec:Eig_th}
 
\begin{figure}[bh]
\begin{center}
    \includegraphics[width=1.\textwidth]{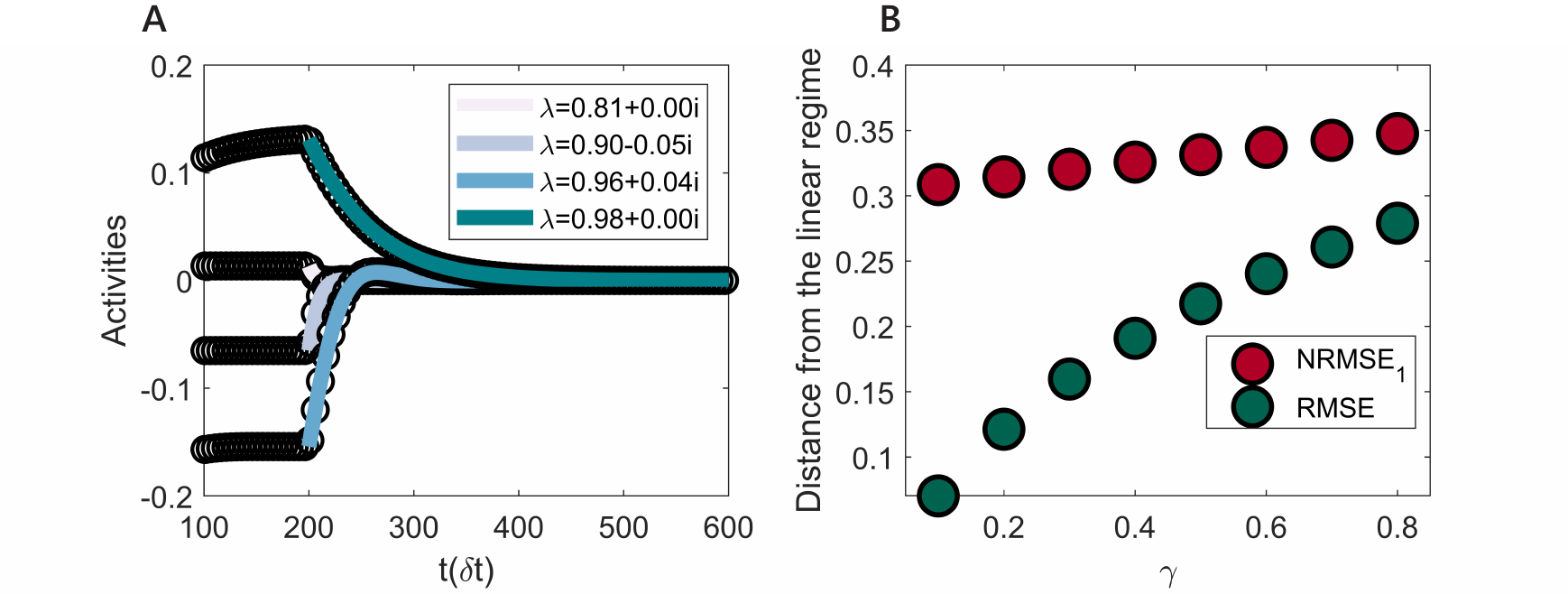}
    \caption{\small{The analysis of the eigenvalues in the linear regime can offer insights in the understanding of the dynamical behaviour of the network. \textbf{A}: Experimental (black dots) and theoretical (coloured lines) response computed though Eq.\ (\ref{eq:act_Th}). The timescales estimated are reflected in the exponentially decaying trends shown, while the oscillations are consequent to the imaginary parts of the eigenvalues. \textbf{B}: RMSE (blue) and NRMSE (red) between the activities of the nodes and the dynamic estimated in the linear regime as $\gamma$, the input strength, varies.     }}
  \label{Fig:Act_eig}
  \end{center}
\end{figure}

 We stimulated the reservoir with a square wave of duration $200 \delta t$ (the time frame of the considered simulation) and analysed the system activity after the impulse to study its relaxation dynamics. Thus, we exploited the fact that, given a system described by $\frac{d\mathbf{x}}{dt}=\mathbf{M}\mathbf{x}(t)$ and where $\mathbf{V}$ are the left eigenvectors of $\mathbf{M}$, i.e
\begin{equation}
\mathbf{V}^T\dfrac{d\mathbf{x}}{dt}=\mathbf{V}^T \mathbf{M} \mathbf{x}(t) = \boldsymbol{\Lambda} \mathbf{x}(t),  \label{eq.x_proj}
\end{equation}
Thus the dynamics of the eigenvectors will be given by
\begin{equation}
    \mathbf{V}^T \mathbf{x}(t)=e^{\boldsymbol{\Lambda} t} \mathbf{V}^T \mathbf{x}(0),
\end{equation}
where $\boldsymbol{\Lambda}$ is the diagonal matrix composed by the eigenvalues of the matrix $\mathbf{M}$. Of course, in the case considered $\mathbf{M}=(1-\alpha)\mathbb{I}+\alpha \mathbf{W}$ and $\text{Re}(\lambda)=1-\alpha + \alpha \lambda_{\mathbf{W}}$, $\text{Im}(\lambda)=\alpha \lambda_{\mathbf{W}}$. Thus, considering a column $\mathbf{v}$ of $\mathbf{V}$ and the corresponding eigenvalue $\lambda$
\begin{equation}
   \mathbf{v}^T\mathbf{x}(t)=e^{\mathrm{Re}(\lambda) t}\Big[ \mathrm{Re}\big(\mathbf{v}^T\mathbf{x}(0)\big)\cos\big(\mathrm{Im}(\lambda)t\big)- \mathrm{Im}\big(\mathbf{v}^T\mathbf{x}(0)\big)\sin\big(\mathrm{Im}(\lambda)t\big)\Big], \label{eq:act_Th}   
\end{equation}
can be used to compare the true dynamic $\mathbf{V}^T \mathbf{x}(t)$ with the linearised one. Fig.\ \ref{Fig:Act_eig} shows the result of this procedure for each dimension of $\mathbf{V}^T \mathbf{x}(t)$. Panel A reports example activities and their corresponding theoretical trend for the case of small input values ($\gamma=0.05$, see \ref{eq:input}), case in which the system can be well approximated through a linear behaviour. Panel B shows the $\rm RMSE$ and $\rm NRMSE_1$ between the experimental activities and the theoretical one as $\gamma$ increases. 
In this case, with $y_i(t)=\mathbf{v}^T\mathbf{x}(t)$ experimentally observed, while $\tilde{y}_i(t)$ estimated through the right side of Eq.\ \ref{eq:act_Th}
\begin{equation}
    \rm NRMSE_1=\sum_i \dfrac{1}{N |\max(y_i)-\min(y_i)|}\sum_t \sqrt{\dfrac{\big(\tilde{y}_i(t)-y_i(t)\big)^2}{T-1}} 
\end{equation}

\subsection{Delayed Signal to ESN}

\begin{figure}[h!]
\begin{center}
    \includegraphics[width=0.55\textwidth]{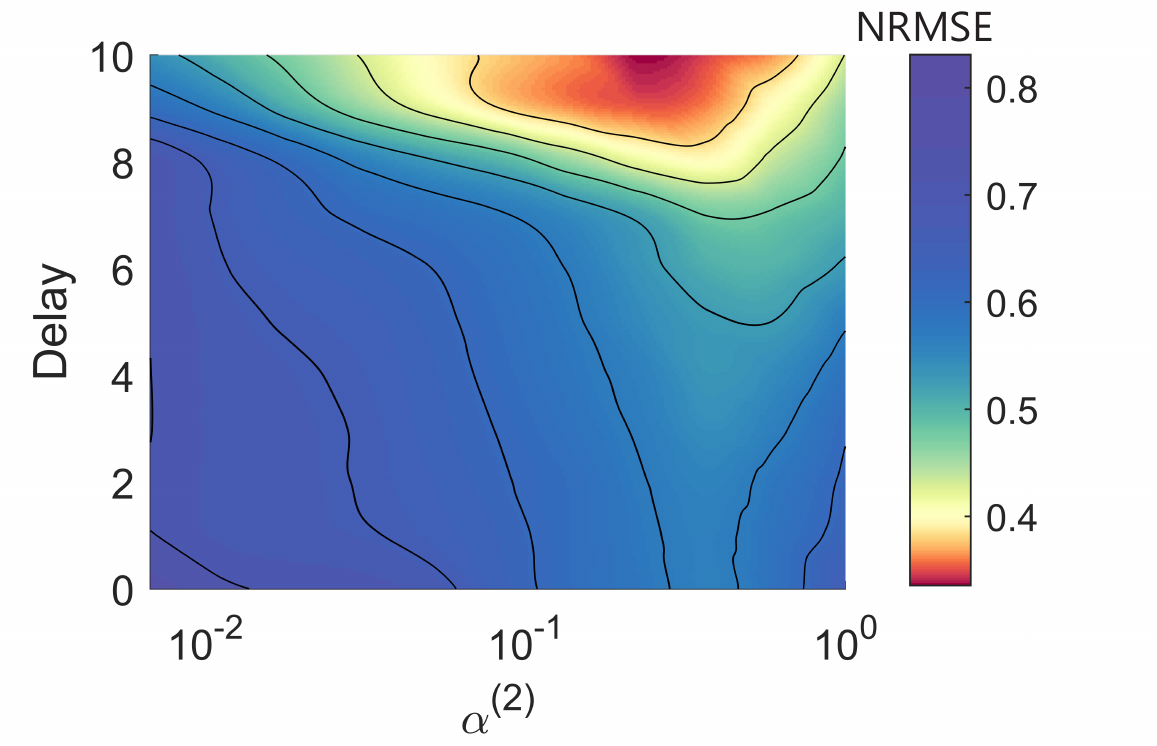}
    \caption{\small{ Performance of an ESN when the signal is expanded by addition of its previous temporal steps in the NARMA10 task. The lowest error corresponds to a leakage term $\alpha^{(2)}$ that is in agreement with the optimal value of $\alpha^{(2)}$ of the connected ESN structure reported Section \ref{Sec:narma} (Main Text). }}
  \label{Fig:Delay_ESN}
  \end{center}
\end{figure}

\new{We computationally validate the equation \ref{Eq:HESN_rep3} (below), derived in Section \ref{Sec.HESN}, on the NARMA10 task. The NARMA10 task is described in full in Section \ref{Sec:narma} (Main Text).}

\begin{equation}
    \dot{\mathbf{y}}^{(2)} = \boldsymbol{\Lambda}^{(2)} \, \mathbf{y}^{(2)} - \tilde{\mathbf{W}}^{(12)} \, \mathbf{T}^{(1)} \, \left[
     \tilde{\mathbf{W}}_{\rm in}^{(1)} \mathbf{s}(t) + \sum_{n=1}^\infty (\mathbf{T}^{(1)})^n \tilde{\mathbf{W}}_{\rm in}^{(1)} \frac{d^{(n)} \mathbf{s}(t) }{ dt^{(n)}} \right]. \label{Eq:HESN_rep3}
\end{equation}

\new{In order to approximate the scaling of the coefficients of the derivatives in Eq.\ \ref{Eq:HESN_rep3}, we incorporate a delay into the input signal such that the activity of the first reservoir is replaced by
\begin{equation}
x^{(1)}_i(t) = \sum_{j=0}^{\mathrm{Delay}} \xi_{ij} \, 0.8^j \, s(t - j)
\end{equation}
where $\xi_{ij}$ are independent Gaussian variables of variance $\sigma^2_\xi$ chosen such that $\mathrm{Var}[x^{(1)}_i] = 1$ for every $i$ and every value of Delay. In practice, we adopted the following approximation;}
\begin{align}
    x_i(t)\cong \mathbf{T}^{(1)} \, \left[
     \tilde{\mathbf{W}}_{\rm in}^{(1)} \mathbf{s}(t) + \sum_{n=1}^\infty (\mathbf{T}^{(1)})^n \tilde{\mathbf{W}}_{\rm in}^{(1)} \frac{d^{(n)} \mathbf{s}(t) }{ dt^{(n)}} \right]. 
\end{align}
The stochastic elements $\xi_{ij}$ emulate the random mixing matrix that, in Eq.\ \ref{Eq:HESN_rep3}, projects the expanded input onto the second reservoir network. 

We compare the result obtained for the hierarchical network, reported in Fig. \ref{fig:narma}G, with the one illustrated in Fig.~\ref{Fig:Delay_ESN}, where the first network has been replaced by Eq.\ \ref{Eq:HESN_rep3}, for different delays (equivalent to different orders of retained derivatives). Figure \ref{Fig:Delay_ESN} shows that as the delay increases, thus higher derivatives are included, the performance appears to converge to an optimal value of $\alpha^{(2)}$ very close to the one in Fig. \ref{fig:narma}G. We also notice that the analysis illustrated earlier suggests that optimal performances are obtained for small $\alpha^{(1)}$. \new{The agreement of results confirms the validity of the approximation used in deriving Eq.\ \ref{Eq:HESN_rep3}.}\newline \vspace{1em}

\bibliographystyle{unsrt}
\bibliography{references}

%\section{Examples of citations, figures, tables, references}
%\label{sec:others}
%\lipsum[8] \cite{kour2014real,kour2014fast} and see \cite{hadash2018estimate}.

%The documentation for \verb+natbib+ may be found at
%\begin{center}
%  \url{http://mirrors.ctan.org/macros/latex/contrib/natbib/natnotes.pdf}
%\end{center}
%Of note is the command \verb+\citet+, which produces citations
%appropriate for use in inline text.  For example,
%\begin{verbatim}
%   \citet{hasselmo} investigated\dots
%\end{verbatim}
%produces
%\begin{quote}
%  Hasselmo, et al.\ (1995) investigated\dots
%\end{quote}

%\begin{center}
%  \url{https://www.ctan.org/pkg/booktabs}
%\end{center}

%\subsection{Figures}
%\lipsum[10] 
%See Figure \ref{fig:fig1}. Here is how you add footnotes. \footnote{Sample of the first footnote.}
%\lipsum[11] 

%\begin{thebibliography}{1}

%\bibitem{kour2014real}
%George Kour and Raid Saabne.
%\newblock Real-time segmentation of on-line handwritten arabic script.
%\newblock In {\em Frontiers in Handwriting Recognition (ICFHR), 2014 14th
%  International Conference on}, pages 417--422. IEEE, 2014.

%\bibitem{kour2014fast}
%George Kour and Raid Saabne.
%\newblock Fast classification of handwritten on-line arabic characters.
%\newblock In {\em Soft Computing and Pattern Recognition (SoCPaR), 2014 6th
%  International Conference of}, pages 312--318. IEEE, 2014.

%\bibitem{hadash2018estimate}
%Guy Hadash, Einat Kermany, Boaz Carmeli, Ofer Lavi, George Kour, and Alon
%  Jacovi.
%\newblock Estimate and replace: A novel approach to integrating deep neural
%  networks with existing applications.
%\newblock {\em arXiv preprint arXiv:1804.09028}, 2018.

%\end{thebibliography}

\end{document}